\documentclass[lettersize,journal]{IEEEtran}
\usepackage{amsmath,amsfonts}
\usepackage{algorithmic}
\usepackage{algorithm}
\usepackage{array}
\usepackage[caption=false,font=normalsize,labelfont=sf,textfont=sf]{subfig}
\usepackage{textcomp}
\usepackage{stfloats}
\usepackage{url}
\usepackage{verbatim}
\usepackage{graphicx}
\usepackage{cite}
\usepackage{booktabs}
\usepackage{mathrsfs}%
\usepackage{textcomp}%
\usepackage{manyfoot}%

\usepackage[acronym]{glossaries}
\usepackage{csquotes,xpatch}
\usepackage{lscape}
\usepackage{threeparttable}
\usepackage{adjustbox}

\newacronym{3dod}{3DOD}{3D Object Detection}
\newacronym{2dod}{2DOD}{2D Object Detection}
\newacronym{uw3dod}{Underwater-3DOD}{Underwater 3D Object Detection}
\newacronym{sota}{SoTA}{State-of-the-Art}
\newacronym{roi}{ROI}{Region of Interest}
\newacronym{tof}{ToF}{Time-of-Flight}
\newacronym{pcl}{PCL}{Point Cloud}
\newacronym{cnn}{CNN}{Convolutional Neural Networks}
\newacronym{ann}{ANN}{Artificial Neural Network}
\newacronym{bev}{BEV}{Bird's Eye View}
\newacronym{icp}{ICP}{Iterative Closest Point}
\newacronym{cad}{CAD}{ Computer-Aided Design}
\newacronym{ransac}{RANSAC}{Random Sample Consensus}
\newacronym{pfh}{PFH}{Point Feature Histograms}
\newacronym{pca}{PCA}{Principal Component Analysis}
\newacronym{mlp}{MLP}{Multiplayer Perception Network}
\newacronym{imr}{IMR}{Inspection, Maintenance and Repair}
\newacronym{mbes}{MBES}{Multibeam Echosounder}
\newacronym{dol}{DOL}{Digital Ocean Lab}
\newacronym{dbscan}{DBSCAN}{Density-Based Spatial Clustering}
\newacronym{iou}{IoU}{Intersection over Union}
\newacronym{map}{mAP}{Mean Average Precision}
\newacronym{ap}{AP}{Average Precision}

\graphicspath{images}
\begin{document}
\bstctlcite{bstctl:nodash}

\title{Towards Training-Free Underwater 3D Object Detection from Sonar Point Clouds: A Comparison of Traditional and Deep Learning Approaches}

\author{
M. Salman Shaukat\textsuperscript{1,*}, Yannik Käckenmeister\textsuperscript{1}, Sebastian Bader\textsuperscript{1}, and Thomas Kirste\textsuperscript{1}
\thanks{\textsuperscript{1}Computer Science Department, University of Rostock, Rostock, Germany.}
\thanks{*Corresponding author: muhammad.shaukat@uni-rostock.de}
}



\maketitle
\begin{abstract}
Underwater 3D object detection remains one of the most challenging frontiers in computer vision, where traditional approaches struggle with the harsh acoustic environment and scarcity of training data. While deep learning has revolutionized terrestrial 3D detection, its application underwater faces a critical bottleneck: obtaining sufficient annotated sonar data is prohibitively expensive and logistically complex, often requiring specialized vessels, expert surveyors, and favorable weather conditions. 
This work addresses a fundamental question: Can we achieve reliable underwater 3D object detection without real-world training data? We tackle this challenge by developing and comparing two paradigms for training-free detection of artificial structures in multibeam echo-sounder point clouds. Our dual approach combines a physics-based sonar simulation pipeline that generates synthetic training data for state-of-the-art neural networks, with a robust model-based template matching system that leverages geometric priors of target objects.
Evaluation on real bathymetry surveys from the Baltic Sea reveals surprising insights: while neural networks trained on synthetic data achieve 98\% mean Average Precision (mAP) on simulated scenes, they drop to 40\% mAP on real sonar data due to domain shift. Conversely, our template matching approach maintains 83\% mAP on real data without requiring any training, demonstrating remarkable robustness to acoustic noise and environmental variations.
Our findings challenge conventional wisdom about data-hungry deep learning in underwater domains and establish the first large-scale benchmark for training-free underwater 3D detection. This work opens new possibilities for autonomous underwater vehicle navigation, marine archaeology, and offshore infrastructure monitoring in data-scarce environments where traditional machine learning approaches fail.
\end{abstract}


\section{Introduction \& Background}
The underwater world is both important and challenging for visual computing. Studying and monitoring these environments matters for ecological as well as industrial reasons. They include a wide range of marine ecosystems such as coral reefs, fish populations, and microorganisms. Alongside natural habitats, there are also many human-made structures beneath the ocean surface, including offshore oil and gas platforms, subsea pipelines, communication cables, and artificial reefs.

This research focuses on the challenge of classifying and localizing artificial underwater structures. A key factor in this task is the choice of sensor modality. As discussed in Section~\ref{sec:env_sensors}, sonar is the natural option for underwater exploration and, in deep oceans, often the only practical one. Sonar works by using acoustic reflections to measure depth. In particular, \gls{mbes} systems capture dense sets of 3D points that represent the geometry of the seafloor and any objects present on it.

In computer vision, this type of data is commonly referred to as a 3D point cloud.  
It requires specialized computational methods for analysis and interpretation. Many existing approaches down-sample 3D point cloud data into 2D images, allowing the use of well-established image processing techniques and deep neural networks.  
However, this down-sampling inevitably leads to information loss.  In our work, we remain in the 3D domain to fully exploit the richness of point cloud data. This allows us to estimate the size, position, and orientation of objects, rather than only performing classification. As a result, our method addresses a \gls{3dod} task.  

Regarding the \gls{3dod} field, substantial research has been carried out, especially in the deep learning domain. This is mainly driven by the growing interest in autonomous driving, which relies on LiDAR sensors that produce 3D point clouds of the vehicle's surroundings for navigation. As expected, the performance of deep learning-based \gls{3dod} methods depends heavily on the availability of large amounts of \textit{annotated} training data.

In underwater environments, however, obtaining training data is extremely difficult due to several factors. It requires careful planning by expert surveyors and involves costly resources such as specialized equipment and personnel \cite{xie2022dataset}. Unlike terrestrial or aerial data collection, the extreme underwater environment presents unique challenges. Weather conditions and environmental factors make the process complex and prone to errors \cite{er2023research}. Additionally, because of the high cost of data collection and issues of confidentiality, most datasets remain private \cite{aubard2025sonar} and are therefore of limited use for research and development.

In the absence of sonar training data, this work focuses on \textit{training-less} underwater 3D object detection, where real sonar data is used \textit{only} during the testing phase. In particular, we target man-made structures where the shape and size of objects are known as background knowledge.  

Regarding the choice of detection methods for \textit{our} research focus, an obvious direction is a deep learning-based solution trained on simulated sonar data. This choice is motivated by the widespread adoption, rapid development, and strong performance of neural-based approaches in recent years. Therefore, we developed an underwater sonar simulator and used the generated synthetic data to train a state-of-the-art neural network. The trained network was then directly applied to detect objects from real-world sonar data.  

Beyond learning-based approaches, there also exist several \textit{traditional} model-based methods for \gls{3dod}. These involve creating a library of object models and applying them in a template-matching manner. Model-based methods are therefore compelling candidates for training-less \gls{3dod} as well. For this purpose, we constructed a model template library by generating 3D polygon mesh models of objects and converting them into sonar point cloud templates. These model templates were then applied directly to raw sonar data.  

The paper is organized as follows: The remainder of this section introduces key concepts in underwater environments, sonar sensing, and 3D detection methods. Section~\ref{sec:related_work} reviews existing work on \gls{3dod} in the underwater domain. Section~\ref{sec:dataset} provides detailed information on both the synthetic and real sonar data. Sections~\ref{sec:implementation} and \ref{sec:results} describe the implementation details and present the results, respectively. Finally, Section~\ref{sec:discussion} offers a discussion and an outlook on future work.

\subsection{Underwater Environment and Sensors}
\label{sec:env_sensors}

In underwater environments, computer vision application areas include surveillance, scientific exploration, and industrial repair \& maintenance \cite{SAHOO2019145}. Despite its importance, underwater perception faces challenges that are not present in terrestrial or aerial environments. These challenges include light distortion, attenuation, and scattering, which lead to image degradation and loss of contrast \cite{gonz2023}. As a result, optical vision deteriorates with increasing depth, particularly due to the lack of natural light. Therefore, acoustic sonar sensors are widely used in underwater settings because acoustic signals are more robust against attenuation compared to optical sensors such as cameras or LiDAR \cite{HUY2023113202}.  

There are numerous types of underwater sonar sensors in use today, differing in both their design principles and the data they record. They are generally categorized into three types: singlebeam (or forward-looking), multibeam, and sidescan sonar. A multibeam sonar, as the name suggests, emits hundreds of beams on both sides and maps the seafloor in high resolution. In this work, the data was recorded using a similar multibeam sonar towed by a surface vessel. For further reading on sonar sensors, a comprehensive review is provided in \cite{s21237849}.

\subsubsection{Data Recording in the Baltic Sea (Digital Ocean Lab)}
\label{sec:dol}

Between 2019 and 2021, the Mecklenburg-Vorpommern Research Center for Agriculture and Fisheries in Germany initiated the construction of an artificial reef in the Baltic Sea, in collaboration with the European Maritime and Fisheries Fund (EMFF). This project, known as the \enquote{Digital Ocean Lab} (DOL), created underwater habitats that serve as recruitment, growth, and resting zones for local fish species \cite{oceans_fisheries_2023, reef_nienhagen_2022}.  

\begin{figure*}[!t]
  \centering
  \includegraphics[width=\textwidth]{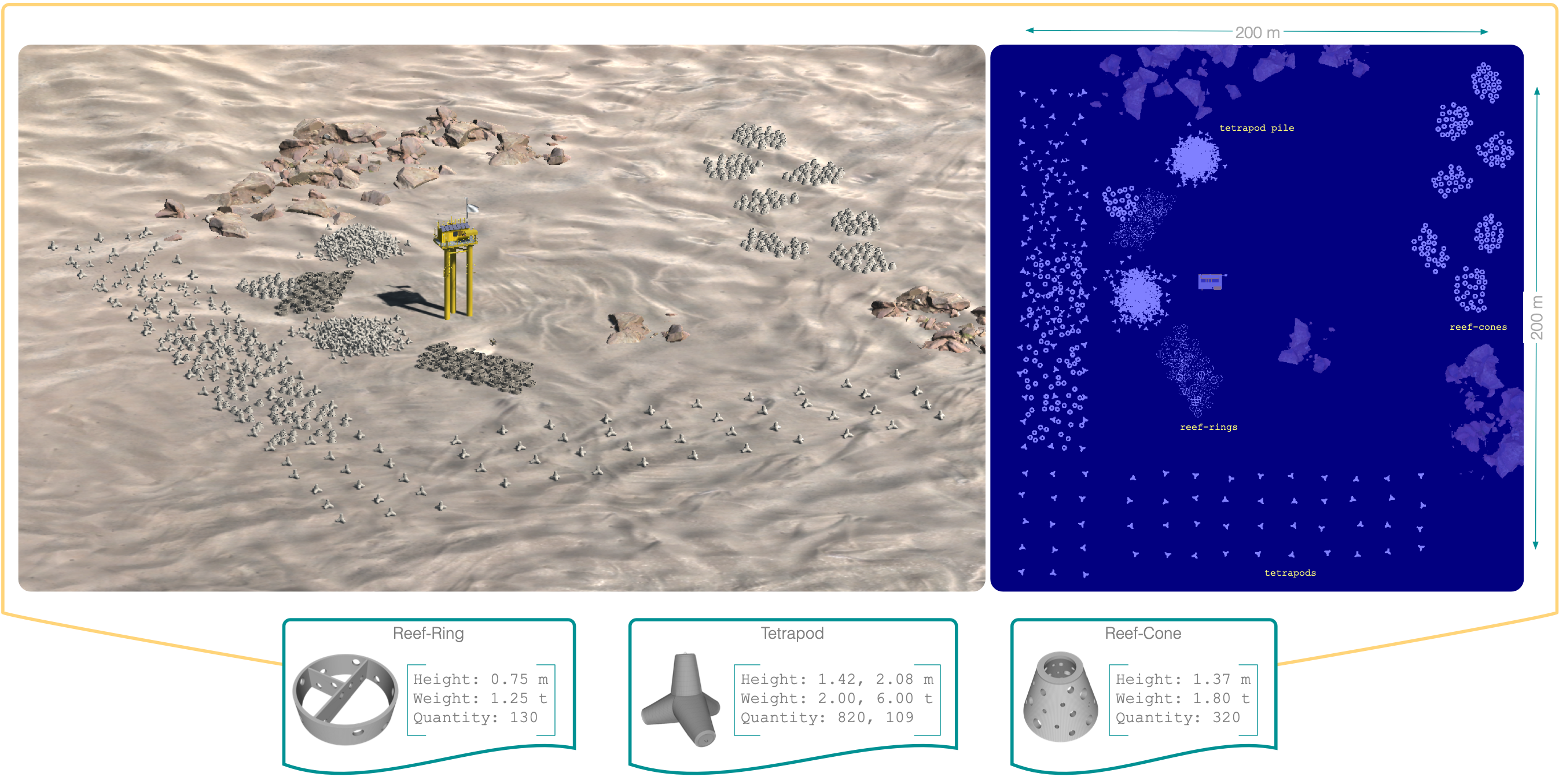}
  \caption{Digital Ocean Lab (DOL) overview. (Left) Side view showing underwater placement. (Right) Top view displaying spatial distribution of artificial reef structures over 200×200 meter area. 
  For more information, we refer to \texttt{https://www.riff-nienhagen.de}.}
  \label{fig:dol}
\end{figure*}

The DOL testing site includes man-made concrete structures such as wave-dissipating blocks (tetrapods), reef\_rings, and reef\_cones, in addition to natural rocks. Figure~\ref{fig:dol} shows a conceptual 3D depiction of the DOL along with 3D models of the objects present there. The total area of interest is approximately $200 \times 200$ meters and contains nearly 1,400 objects, including \textit{reef\_ring}, \textit{tetrapod}, and \textit{reef\_cone} types. Further details on the recorded sonar data are provided in Section~\ref{sec:sonar_data}.

\subsection{Point Cloud Data}
\label{sec:pcd}

Point cloud data is one of the most fundamental forms of 3D data, representing objects or scenes as a collection of discrete points in a 3D coordinate space. Each point captures a location in space and, depending on the sensing device, may also include attributes such as color, reflectance, or intensity \cite{griffiths2019review, wang2020overview}. In our case, we work with raw, uncolored sonar point cloud data. This lack of additional features makes the detection task inherently more challenging.  

While depth information in point cloud data offers clear advantages, it also introduces challenges unique to 3D representation. Compared to 2D images, point clouds have several difficult characteristics. They are irregular due to uneven point distribution, unstructured because they do not follow a fixed grid, and permutation-invariant, meaning the order of points does not affect the representation \cite{bello2020deep}. These attributes create significant challenges. They complicate the use of conventional deep learning architectures such as convolutional neural networks (\glspl{cnn}), which assume structured input data.  

To address these issues, intermediate representations such as voxelization or multi-view projections are often used. However, these down-sampling transformations result in a loss of geometric detail. In contrast, methods like PointNet \cite{qi2017pointnet} operate directly on raw point sets, preserving fine-grained spatial information. In this work, recognizing the limitations of down-sampling, we employed methods that process sonar point clouds directly.

\subsection{Overview of 3D Detection Methods}
\label{sec:3d_methods}

Widespread interest in autonomous driving has accelerated the development of deep learning-based approaches for 3D data processing. These methods leverage large annotated datasets and modern hardware to enable efficient 3D processing. Section~\ref{sec:nn_approaches} outlines the techniques used in this work. In contrast, traditional model-fitting approaches remain relevant, especially in scenarios where annotated training data is scarce or unavailable. A basic overview of these model-based techniques is provided in Section~\ref{sec:model_approaches}.  

\subsubsection{Deep Learning Based Approaches}
\label{sec:nn_approaches}

The adoption of deep learning in 3D computer vision has accelerated in recent years, driven by its success across diverse tasks. For example, applications include self-driving vehicles in the aerial domain \cite{zimmer2022survey} and robotic object grasping \cite{CzajewskiKołomyjec+2017+219+237}. Deep learning architectures typically require input data in a regularized format such as images, where each pixel corresponds to a spatial location. This structure enables convolutional and other grid-based processing techniques.  

Similarly, in 3D, point cloud data can be \textit{down-sampled} into a 3D grid, allowing the use of methods such as 3D convolutions. These architectures are referred to as \textit{volumetric} learning methods. However, down-sampling inevitably leads to a loss of geometric detail. To address this, researchers have developed methods that directly process raw point cloud data, known as \textit{pointwise} learning methods.  

In volumetric feature learning, point cloud data is segmented into a structured grid representation such as \textit{voxels}, enabling the application of 3D convolutional operations. While this approach has shown promise, particularly with advances like VoxelNet \cite{zhou2018voxelnet} and sparse convolutions \cite{yan2018second}, it struggles with the inherent sparsity of point clouds and the heavy computational cost of 3D convolutions.  

Pointwise feature learning, on the other hand, processes point clouds directly without regularization, thereby preserving the original geometric information. This approach was pioneered by \textit{PointNet} \cite{qi2017pointnet}, which uses \gls{mlp} for feature learning and has since evolved into more sophisticated architectures. One notable example is \textit{SASA} \cite{chen2022sasasemanticsaugmentedsetabstraction}, which performed exceptionally well on the KITTI 3D object detection benchmark among methods that rely solely on point cloud data \cite{kitti}. Therefore, for the deep learning-based approach to underwater 3D object detection in our work, we employed the SASA neural network.

\subsubsection{Model Fitting Based Approaches}
\label{sec:model_approaches}

In model-fitting-based \gls{3dod}, a model of the \textit{target} object is used to identify its presence in the recorded data. A model can be represented mathematically (e.g., a parametric equation of a sphere) or as a template. Templates may be acquired synthetically (i.e., 3D \gls{cad} models of objects) or naturally using 3D sensors. In practice, templates of different objects, captured from multiple viewpoints, are stored in a template database. This database is then used to search for corresponding objects in the recorded data.  

To detect objects using mathematical models, the \gls{ransac} method can be applied. \gls{ransac} fits a parametric model to the data and identifies the \textit{inlier} points that best conform to the model. The inlier count serves as a measure of how well a set of points in the point cloud matches the target model. \gls{ransac} is most often used for detecting primitive 3D shapes such as planes, spheres, or cylinders \cite{martinez-otzeta2023}. Its strength lies in its simplicity and robustness against noise and outliers \cite{kaiser2019}. In the \gls{3dod} domain, \gls{ransac} is frequently employed as a pre-processing step for plane (ground) removal or segmentation \cite{Xie_2020}. In our case, we used \gls{ransac} to remove the seabed from sonar data as a pre-processing step.  

Template matching methods, in contrast, compute the similarity between stored 3D models in the database and potential object instances in the data. Traditionally, this has been achieved with 3D feature descriptors such as \gls{pfh} \cite{laga20183d}. However, in our use case, sonar data was highly noisy, which made feature descriptors unreliable for robust feature computation. To address this, we adopted a direct alignment strategy using the Iterative Closest Point (ICP) algorithm. ICP is an iterative method that alternates between establishing point correspondences between two point clouds and estimating the rigid transformation that minimizes the distance between them.  

ICP is traditionally used for fine-tuning an initial alignment obtained through feature matching. However, in our case, it proved to be the \textit{only} reliable method for both coarse and fine alignment due to the high noise levels in sonar data. Instead of relying on feature descriptors, we performed object detection by directly aligning template models to segmented regions in the data using ICP. This approach avoids dependence on unstable geometric features in noisy or low-resolution data and instead leverages the global geometric consistency between the model and the scene.  

\section{Related Work}\label{sec:related_work}

In recent years, there has been significant progress in the field of \gls{3dod}. Most research, however, has focused on automation and indoor monitoring. In the underwater domain (\gls{uw3dod}), only a few notable contributions have been reported. This section provides a brief overview of the state of the art in underwater 3D perception. Table \ref{tab:literature} outlines the studies that specifically address underwater 3D object detection. Along with the detection algorithm, we summarize the application domain, experimental setting, target objects, and sensors used.  

Previous work can be grouped into two domains: exploration and \gls{imr}. In exploration, the goal is to observe the state of underwater environments such as coral reefs \cite{s.bhandarkar2022}. In contrast, \gls{imr} focuses on the inspection and maintenance of underwater infrastructure such as plumbing hardware. As shown in Table \ref{tab:literature}, most prior work targets primitive objects (cylinders, cubes, tyres) or industrial components such as valves and pipes. Regarding algorithms, deep learning has dominated in recent years, in line with the general trend in computer vision. Nonetheless, feature descriptor and \gls{ransac}-based methods remain relevant due to their advantages in situations where training data is unavailable.  

In the deep learning domain, Abadal et al. \cite{martin-abadal2020, martin-abadal2022} collected an underwater dataset in a controlled pool setting to train PointNet \cite{qi2017pointnet} for detecting pipes and valves. Their model was later tested on sea data. They annotated 262 point clouds in the pool and 22 in the sea over diverse seabeds such as rocks and algae. Although their target objects were relatively simple, the results demonstrated the strong potential of neural networks when sufficient, high-quality training data is available. Similarly, Bhandarkar et al. \cite{s.bhandarkar2022} used $\simeq 5000$ images to reconstruct a 3D mesh of coral reefs and detect coral categories. Tsai et al. \cite{tsai2021} applied \gls{ransac} to remove the seabed plane from point clouds before using PointNet to detect tyres placed on the seabed. Wang et al. \cite{wang2020w} combined a 2D detection step with 3D processing: after extracting object-specific point clouds, they refined alignment and estimated object poses using \gls{icp}.  

While deep learning methods are the obvious choice when large amounts of training data are available, traditional model-fitting approaches remain valuable in its absence. This is especially relevant underwater, where acquiring annotated training data is particularly challenging. Himri et al. \cite{himri2018, himri2018b, himri2019, himri2021} investigated model-fitting approaches using an underwater laser scanner to capture plumbing items (e.g., pipes and valves) in a pool setting. For instance, in \cite{himri2021}, \gls{ransac} was applied both in pre-processing and detection: first to remove the pool surface, and later to detect pipes due to their primitive shape. More complex components, such as butterfly valves, were then detected using template-based methods.

\begin{table*}[!t]
\caption{Related work for underwater 3d object detection showing the application domain, types of objects detected along with sensors and algorithms used.}
\resizebox{\textwidth}{!}{%
\begin{tabular}{@{}llllllll@{}}
\hline
Year & Domain &  Setting         & Object Type & Objects                  & Sensor                    & Algorithm           &                                   \\ \hline
2022 & IMR *                & Real and Controlled         & Plumbing    & Pipes, Valves            & Stereo Camera             & Deep Learning       & \cite{martin-abadal2022} \\
2022 & Exploration        & Real                        & Misc        & Shipwreck, Pipe, Seabed  & Laser                     & Model Fitting & \cite{jungkyungmin2022}  \\
2022 & Exploration        & Real                        & Seabed      & Coral Reef               & Stereo Camera             & Deep Learning       & \cite{s.bhandarkar2022}  \\
2022 & Exploration        & Real                        & Seabed      & Seafloor Plume           & Multibeam Sonar           & Model Fitting & \cite{xinren2022}        \\
2022 & Exploration        & Real                        & Primitives  & Cube                     & LiDAR                     & -                   & \cite{dandi2022}         \\
2022 & IMR                & Real and Simulated          & Plumbing    & Pipes                    & Laser and Camera          & RANSAC              & \cite{amossmith2022}     \\
2021 & IMR                & Controlled                  & Plumbing    & Plane,  Pipes, Valves    & Laser                     & Model Fitting & \cite{himri2021}         \\
2021 & Exploration        & Real                        & Primitives  & Tyre                     & Mechanical Scanning Sonar & Deep Learning       & \cite{tsai2021}          \\
2020 & IMR                & Real and Controlled         & Plumbing    & Pipes, Valves            & Stereo Camera             & Deep Learning       & \cite{martin-abadal2020} \\
2020 & Exploration        & Controlled                  & Misc        & Misc                     & Stereo Camera             & Deep Learning       & \cite{wang2020w}         \\
2019 & IMR                & Controlled   and  Simulated & Plumbing    & Pipes, Valves            & Laser                     & Model Fitting & \cite{himri2019}         \\
2018 & IMR                & Controlled                  & Plumbing    & Pipes, Valves            & Laser                     & Model Fitting & \cite{himri2018}         \\
2018 & IMR                & Controlled                  & Plumbing    & Pipes, Valves            & Laser                     & Model Fitting & \cite{himri2018b}        \\
2015 & IMR                & Real                        & Primitives  & Cylinder,   Cube, Sphere & Other                     & Model Fitting & \cite{aggarwal2015a}     \\
2014 & IMR                & Real                        & Plumbing    & Pipes                    & Stereo Camera             & RANSAC              & \cite{fabio2014}         \\
2014 & IMR                & Real                        & Plumbing    & Pipes                    & Stereo Camera             & RANSAC              & \cite{fabjankallasi2014} \\
2005 & IMR                & Real                        & Primitives  & Cylinder                 & Other                     & Hough Transform     & \cite{patel2005}         \\ 

\hline
\end{tabular}%
}
\label{tab:literature}
\vspace{1mm}

\noindent{\footnotesize{\textsuperscript{*}(IMR) Inspection, Maintenance \& Repair.}}

\end{table*}   

\section{Dataset}\label{sec:dataset}

Although our main goal is to analyze raw \gls{mbes} data, we also generate and report synthetic sonar data. The simulated data serves two objectives: i) training the neural network, and ii) enabling a comparison between model-based and neural-based approaches. We first describe the acquisition and specifications of the sonar data, followed by the synthetic data.  

\subsection{Sonar Data}
\label{sec:sonar_data}
\begin{figure}[tb]
    \captionsetup{justification=centering}
    \centering
    \includegraphics[width=0.35\textwidth]{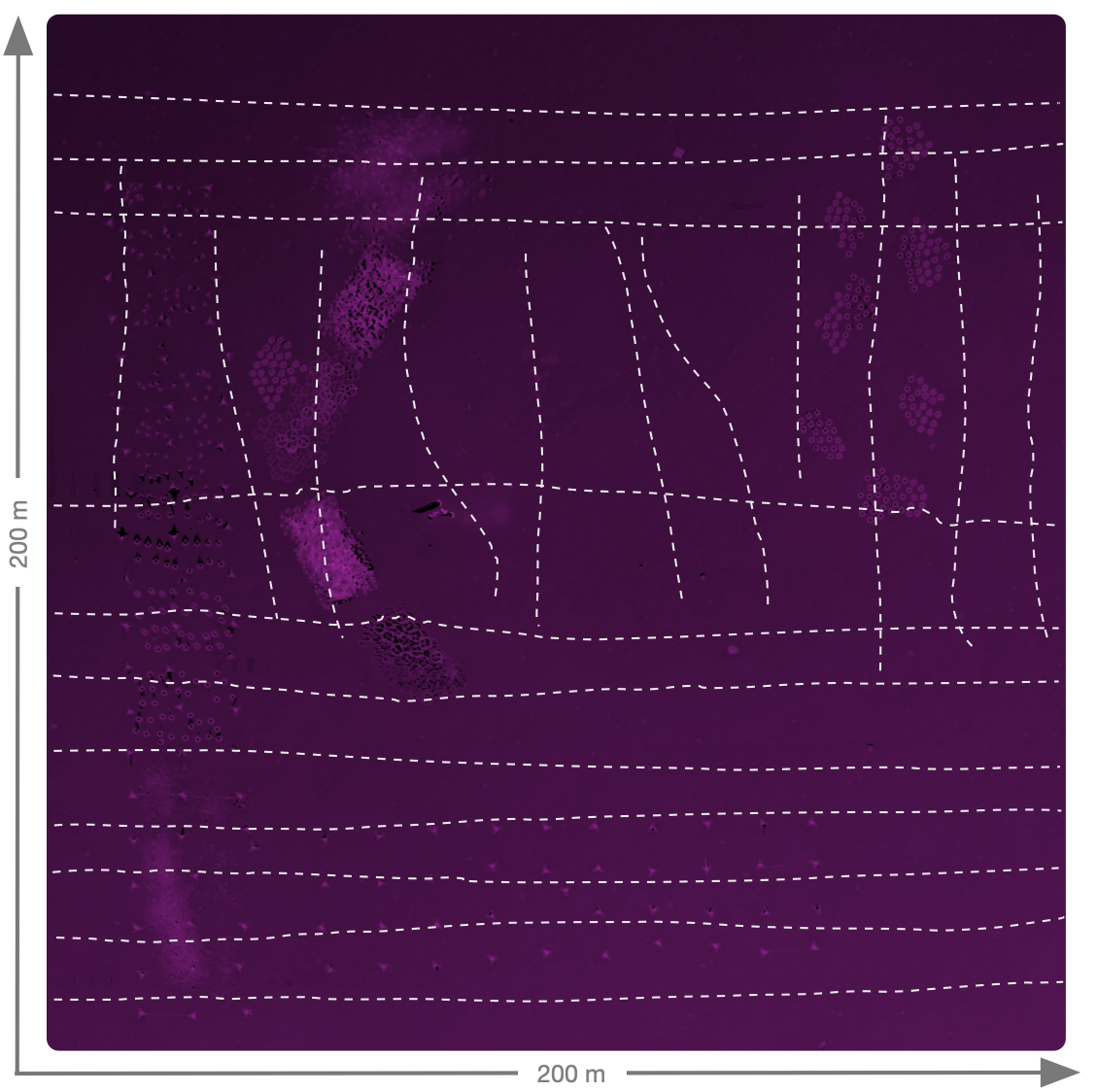}
    \caption{Trajectory of the surface vessel for data recording using an \gls{mbes} sonar.}
    \label{fig:sonar_recording}
\end{figure}

The sonar data was collected by bathymetric mapping of the \gls{dol} test field in the Baltic Sea (Section \ref{sec:dol}) using a \gls{mbes} mounted on a surface vessel. The survey area was mapped over $n=67$ vessel trajectories. Figure \ref{fig:sonar_recording} shows the trajectory lines drawn on top of the acquired sonar data. In total, the sonar scans cover an area of approximately 200 square meters.

\subsubsection{Sonar Data Preprocessing}
\label{subsub:sonar_preprocessing}

In order to feed the sonar data to processing algorithms, we performed a few pre-processing steps. Firstly, to reduce dataset size, we partitioned the data into $100$ equally sized segments. After excluding empty regions (since the covered area is not a perfect square), we obtained $88$ segments.  Furthermore, because most of the covered area did not contain objects of interest, we manually removed the empty segments. In the end, we retained $29$ point clouds, each approximately $46 \times 40$ meters in size. For this work, as a proof-of-concept and evaluation, we annotated $15$ of these segments. Hence, throughout this work, we use these $15$ sonar samples to report our findings. 

\begin{figure*}[b]
    \captionsetup{justification=centering}
    \subfloat[]{
        \includegraphics[width=0.3\textwidth, trim={15cm 1cm 15cm 0cm}, clip]{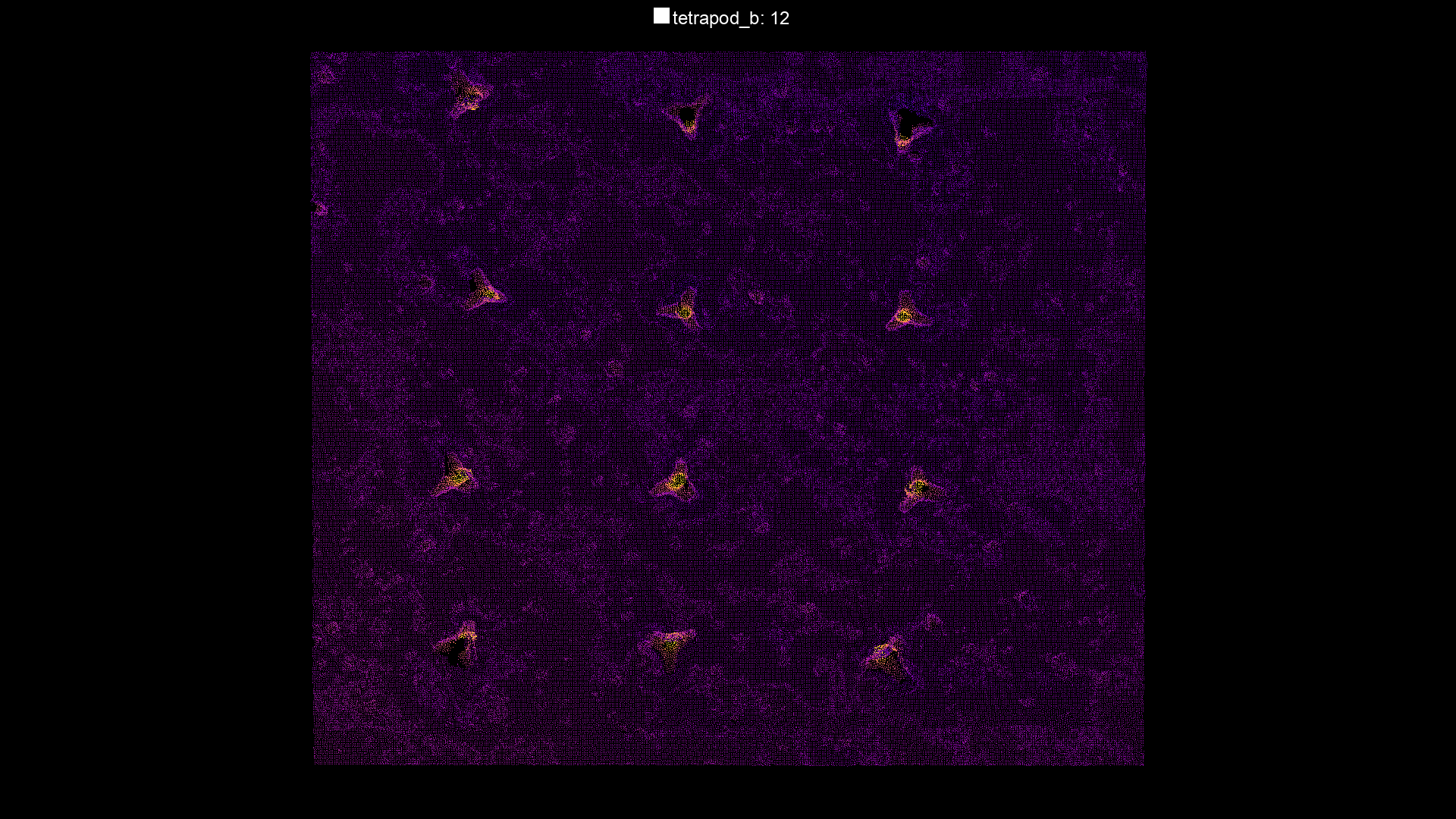}
    }
    \hfill
    \subfloat[]{
        \includegraphics[width=0.30\textwidth, trim={15cm 1cm 15cm 0cm}, clip]{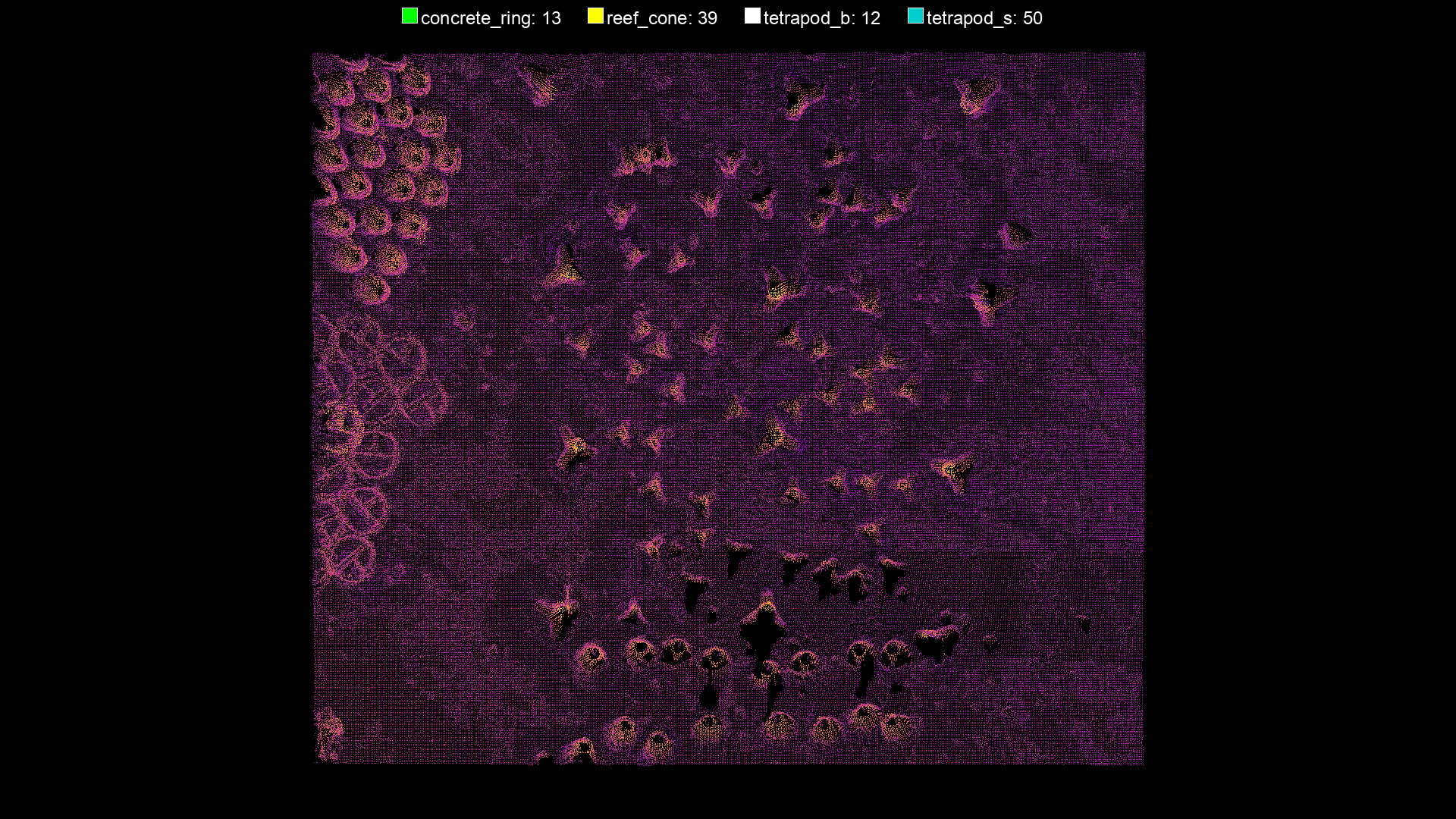}
    }
    \hfill
    \subfloat[]{
        \includegraphics[width=0.3\textwidth, trim={15cm 1cm 15cm 0cm}, clip]{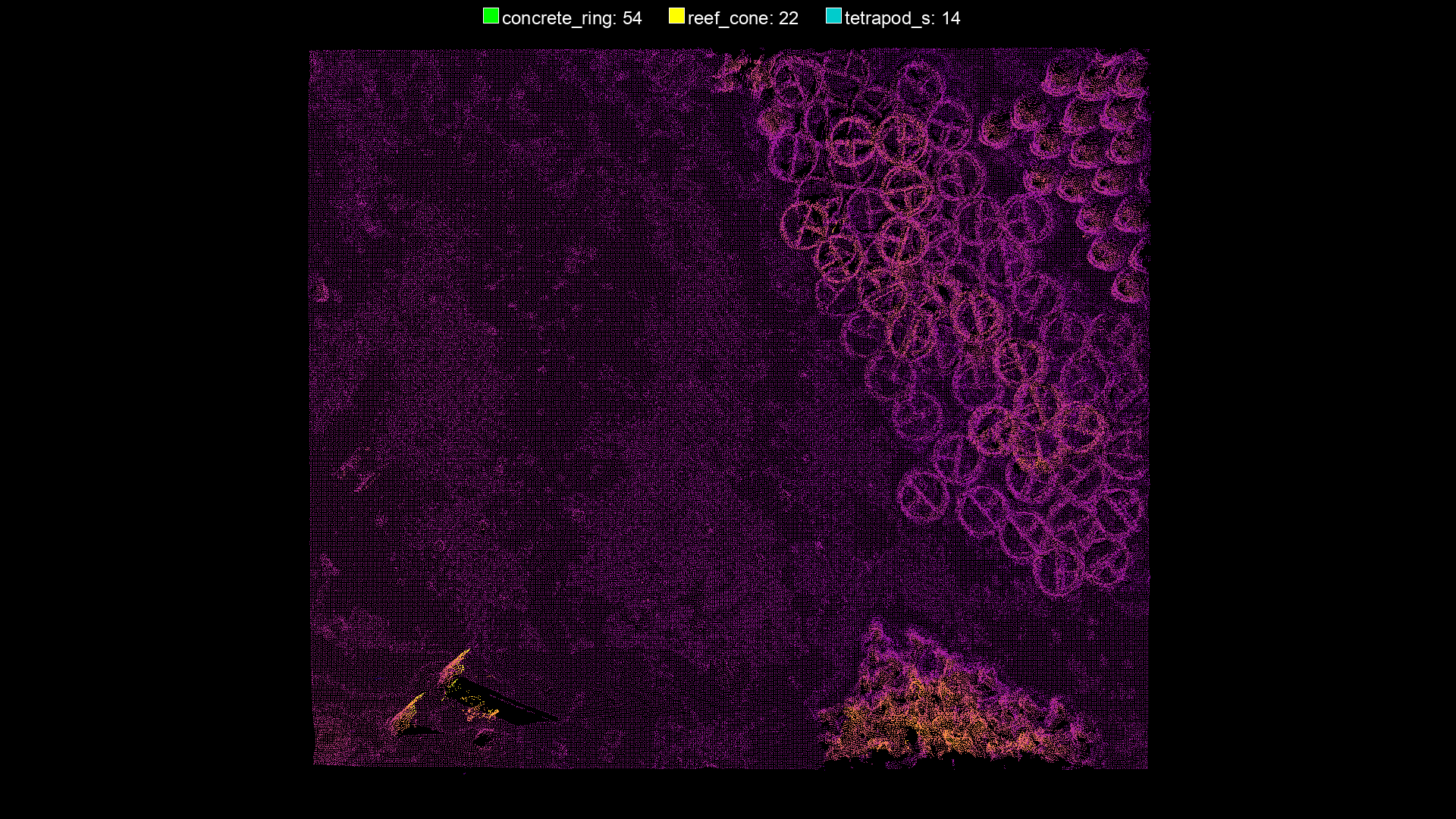}
    }
    \caption{Examples of \gls{mbes} sonar point cloud segments with different complexity levels}
    \label{fig:sonar_images_side_by_side}
\end{figure*}

Figure \ref{fig:sonar_images_side_by_side} shows the top view of three such sonar samples. As seen, the samples exhibit varying levels of complexity. In the first sample, $12$ \textit{tetrapod\_b} (larger tetrapods) objects are present, each positioned with sufficient spatial separation from the others. The second sample contains objects of all types, with noticeably reduced spatial separation. Finally, the last sample shows a more challenging case with clutter and piles of \textit{reef\_ring} objects. We used these three samples for visual reporting of our results, while Table \ref{tab:dataset_stats} outlines the combined object distribution of all $15$ annotated samples.  

\subsection{Synthetic Data}

Sonar data synthesis requires two key components: (a) creating realistic underwater environments and (b) simulating sonar sensors to sample data points. The underwater environment should contain a realistic seabed with both natural and artificial (our domain) objects placed in a physically plausible way. Moreover, to generate a diverse and automated sonar dataset for neural network training, the environment generation must be procedural. This ensures that each data sample exhibits varying seabed characteristics and object placement.  The sonar sensor simulation requires: (i) modeling the spatial movement of the sensor as it scans the environment, and (ii) reproducing the multi-beam acoustic geometry of the sonar, which defines the points sampled from the environment.  

\begin{figure*}[tb]
    \captionsetup{justification=centering}
    \centering
    \includegraphics[width=0.8\textwidth]{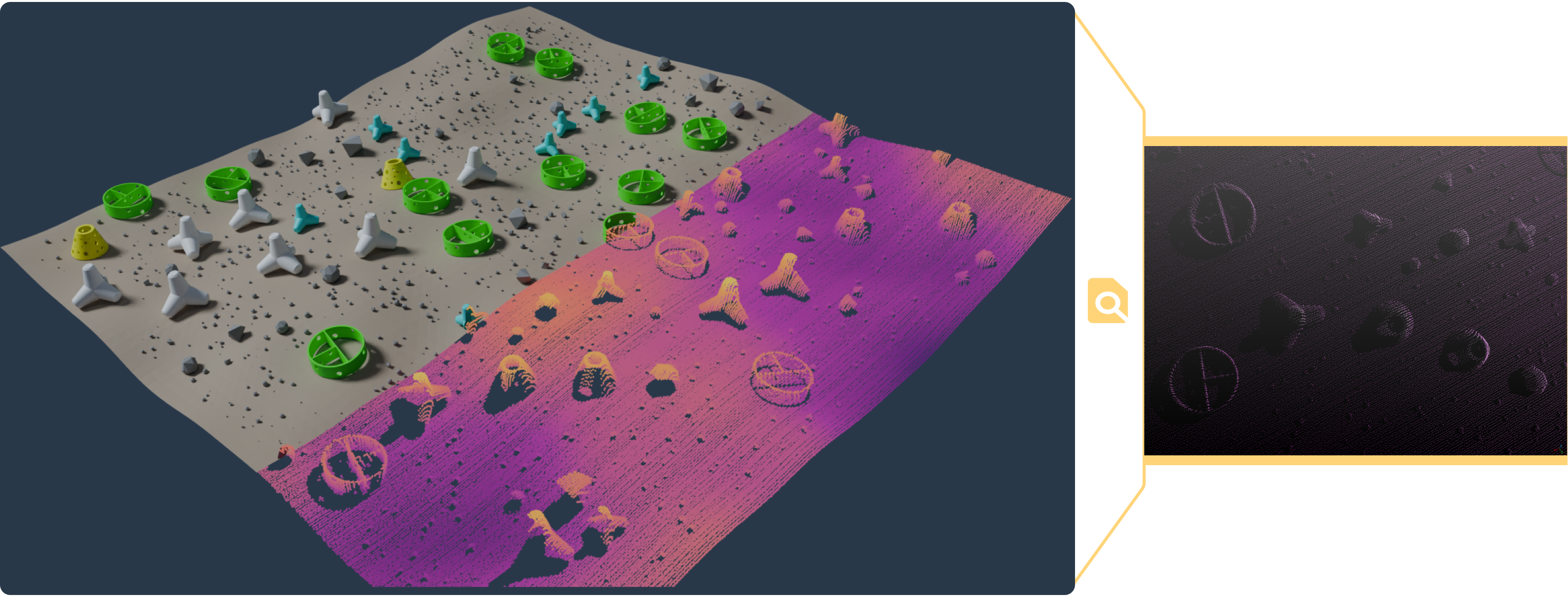}
    \caption{\gls{mbes} Simulation Process: (left) an instance of the simulated 3D environment with seabed, objects and rocks also showing the \gls{mbes} scanning process from above and (right) zoomed in section of the generated point cloud data.}
    \label{fig:blender}
\end{figure*}

\subsubsection{Seabed Generation and Object Placement}

To train neural networks, numerous randomized yet plausible data samples are required. Hence, the process must be procedural with minimal supervision. Seabed generation and object placement are therefore performed in a fully automated manner.  

For seabed generation, we use Perlin noise \cite{perlin1985image}, a gradient noise function that produces smooth and continuous random values. This results in organic-looking terrain with varying levels of roughness.  Once a terrain is generated, objects are placed on it. They are initially positioned at a configurable height above the seabed and then allowed to fall under gravity, with their final positions determined by physics simulation. This creates natural-looking arrangements where objects settle in physically plausible positions on the seafloor.

\subsubsection{Multi-beam Sonar Simulation}

Once the synthetic 3D scene with seabed and objects is created, a sonar scan can be simulated. 
We model an \gls{mbes} sensor traversing above the 3D environment, imitating a surface vessel at a configurable height. 
The virtual sensor follows a predefined trajectory, and at each step it emits a fan of acoustic beams modeled using ray casting. 
Each beam’s return is computed by intersecting rays with the 3D scene geometry.  

The number of beams is configurable, but to match the real \gls{mbes} used in our domain, we fix it to a maximum of $n=1024$ beams. 
Noise fluctuations are included to simulate beams with no returns. 
To further model sensor noise, we characterized the sonar using a flat algae-covered table from our survey data, where the ground-truth geometry (a plane) was known. Point-to-plane distance analysis revealed near-zero bias ($\mu \approx -0.001$\,m) and a standard deviation of $\sigma \approx 0.01$\,m. Accordingly, we approximate the measurement noise using a Gaussian distribution: $\mathcal{N}(0, 0.01^2)$. The detailed methodology and normality assessment are provided in the Appendix.  

The ray casting approach allows simulation of multiple beams at each sensor position while only recording areas within the line of sight. This naturally accounts for partial self-occlusion, as objects can block beams and create sonar shadows.  

Finally, the movement direction of the virtual sensor must be specified, since different trajectories produce different point distributions. In the real survey (section \ref{sec:sonar_data}), mapping was performed mainly along the $x$- or $y$-axis, with some areas recorded in both directions, resulting in denser point clouds. To replicate this effect, our simulation randomly selects the scanning direction to be along the $x$-axis, $y$-axis, or both, thereby producing synthetic point distributions that closely resemble real sonar data.

\subsection{Synthetic Dataset}
Using our simulation tool, we created 100 synthetic \gls{mbes} sonar samples with a total of 5491 objects. We used 80 scenes for training the neural network and 20 for testing. Same testing set was also used to evaluate the model-based approach. As shown in Table \ref{tab:dataset_stats}, both training and test sets maintain a nearly equal distribution of object classes, resulting in a representative dataset. 
Figure~\ref{fig:blender} shows an example of the created synthetic environment and the simulated \gls{mbes} sonar point cloud from it.
\begin{table}[tb]
\centering
\caption{Dataset statistics for synthetic and real sonar data}
\label{tab:dataset_stats}
\resizebox{\columnwidth}{!}{%
\begin{tabular}{lcccccc}
\hline
Dataset & Scenes & Total Objects & Reef\_Ring & Reef\_Cone & Tetrapod\_B & Tetrapod\_S \\
\hline
Synthetic (Train) & 80 & 4518 & 1144 (25.3\%) & 1108 (24.5\%) & 1148 (25.4\%) & 1118 (24.7\%) \\
Synthetic (Test) & 20 & 973 & 236 (24.3\%) & 255 (26.2\%) & 241 (24.8\%) & 241 (24.8\%) \\
\addlinespace[0.5em]
Sonar (Test) & 15 & 373 & 70 (18.77\%) & 158 (42.36\%) & 81 (21.72\%) & 64 (17.16\%) \\
\hline
\end{tabular}%
}
\end{table}



\section{Implementation}\label{sec:implementation}
We implemented and evaluated two paradigms of 3D object detection methods while maintaining our focal restriction of not using any real training data. The fundamentals of i) model fitting and ii) deep learning  approach are introduced in the section \ref{sec:3d_methods}. Following we outline implementation details of these approaches. 

\subsection{Model Fitting}

\begin{figure*}[tb]
    \captionsetup{justification=centering}
    \centering
    \includegraphics[width=\textwidth]{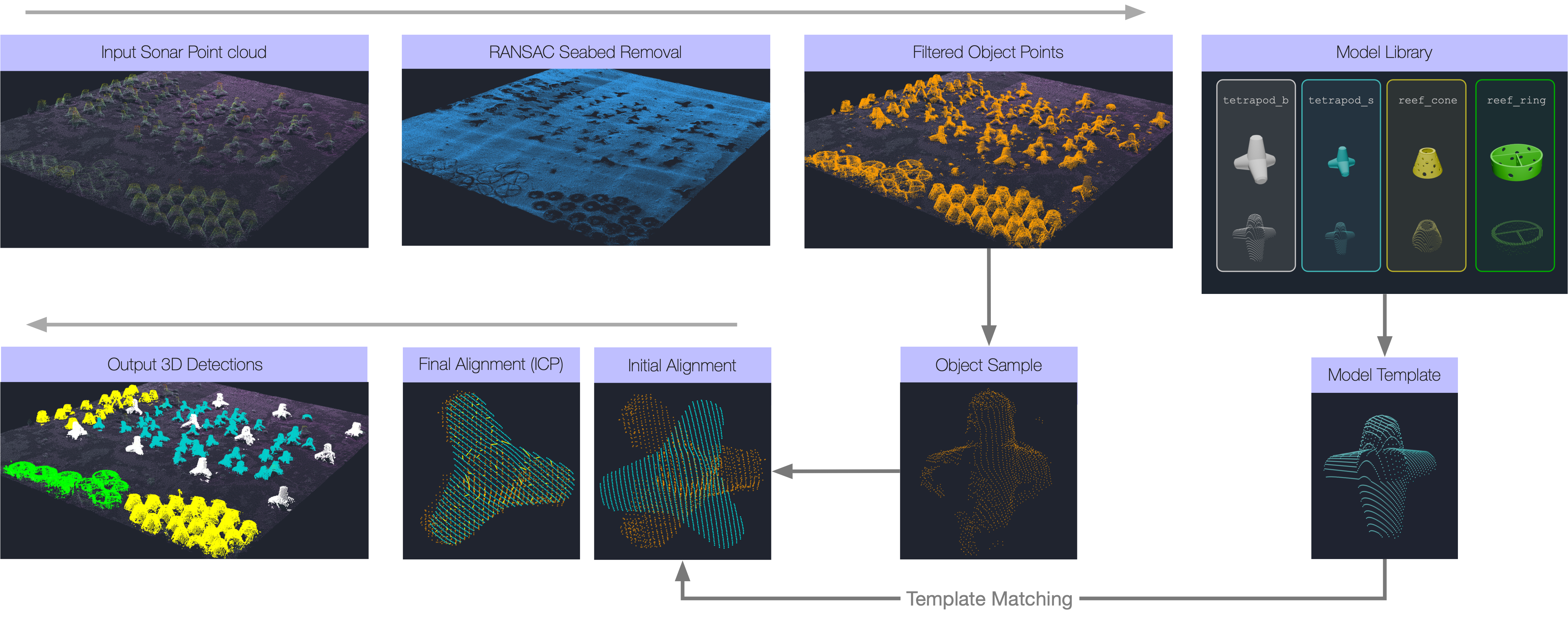}
    \caption{Model-Based Object Detection Pipeline}
    \label{fig:pipeline}
\end{figure*}

Our model fitting approach takes raw sonar point clouds as input and uses object templates to detect their position and orientation in the data as output. The pipeline comprises the following stages:

\subsubsection{Input \& Preprocessing}

Our model-fitting pipeline takes three main inputs: (a) raw sonar point cloud samples, (b) mesh models of the target objects, and (c) a configuration file containing hyper-parameters for subsequent steps.  

The preparation of raw sonar data was described in Section~\ref{sec:sonar_data}, where the complete dataset was partitioned into smaller, more manageable samples ($40 \times 40$\,m). In the \textbf{preprocessing} stage, each sonar sample undergoes seabed removal. A plane is fitted using the \gls{ransac} algorithm, which significantly reduces the point cloud size without losing useful object information. The resulting sample contains points corresponding to potential objects along with some residual seabed noise. This preprocessed point cloud is then passed to the segmentation module.  

For template preparation, mesh models of the target objects are converted into point clouds by synthetically sampling points from their surfaces. To ensure robust matching, the templates must replicate the point distributions produced by real sonar. We therefore simulate the acquisition process by virtually scanning each mesh with a ray-casting approach, analogous to a multibeam echosounder mounted on a surface vessel. This ensures the templates closely resemble the characteristics of the sonar data.

\subsubsection{Segmentation}
After seabed removal, segmentation is applied to extract meaningful regions from the sonar point cloud. This step serves two purposes. First, it eliminates residual noise in areas where point density is too low to represent an object. Second, it generates initial hypotheses indicating potential object locations. Effective segmentation is crucial, as the performance of subsequent detection steps depends heavily on the quality and separation of these candidate segments.  

A major challenge arises in cluttered scenes, where separating overlapping objects remains difficult. Common approaches such as \gls{dbscan} or 3D instance segmentation networks are natural candidates. However, both have limitations. Neural network-based methods require large amounts of annotated data, which is not available in our setting. Density-based methods like \gls{dbscan} rely on clear spatial separation between objects, which cluttered sonar data often lacks.  

To address these challenges, we adopt a simple yet effective sliding-window segmentation. This allows us to focus on object detection rather than complex segmentation design. While this method introduces extra computational cost due to overlapping windows, it preserves object information and provides a reliable basis for detection. The window size is chosen according to the expected object dimensions, and the overlap is defined in the configuration file. Each window segment serves as a model hypothesis for subsequent analysis. To reduce false positives, segments with fewer points than an object-specific threshold are discarded, with the threshold scaled to the size of the target object.  

\subsubsection{Detection and Output}
In the detection step, each candidate segment is evaluated against the object templates generated during preprocessing. For every segment–template pair, an initial alignment is obtained by matching their centroids. This is followed by fine registration using the Iterative Closest Point (\gls{icp}) algorithm. The quality of alignment is measured using the root mean square error (RMSE) between corresponding points after registration. Only matches with an RMSE below an object-specific threshold are retained as valid detections. For each detection, the estimated pose, location, and RMSE score are stored.  

Since overlapping windows can lead to duplicate detections of the same object, a post-processing step is applied to keep only the highest-scoring hypothesis per object. The final output consists of all detected objects, along with their estimated positions and orientations.


\subsection{Deep Learning}\label{sub:deep_learning_methods}

We selected the state-of-the-art SASA (Semantics-Augmented Set Abstraction) neural network for 3D object detection. SASA is designed to work directly with point cloud data and focuses on the most relevant parts of the scene by learning to ignore background points. We adapted its configuration to our dataset and evaluated its performance on sonar data. For detailed technical information about the network, readers are referred to the original publication \cite{chen2022sasa}.

The SASA network was trained on a synthetic dataset of 80 scenes for 100 epochs using an NVIDIA L40 GPU. To improve robustness, standard data augmentation techniques were applied, including scene-level geometric transformations (rotation, scaling, flipping) and object-level bounding box noise injection. Training used the Adam optimizer with an initial learning rate of 0.01, a batch size of 8, and a scheduled learning rate decay to ensure convergence.


\section{Results}\label{sec:results}
Following, we first describe the evaluation method used to compare the, later outlined, results for model fitting and deep learning approach on both synthetic data and sonar data.
\subsection{Evaluation Metric}
\label{sec:metrics}

To evaluate detection performance, we adopt the official nuScenes evaluation framework \cite{caesar2020nuscenes}, which computes the \gls{map} for 3D object detection. The evaluation proceeds in two steps. First, predicted detections are matched to ground-truth objects using predefined center-to-center distance thresholds (i.e., 0.5 meters), capturing spatial alignment between predictions and annotations. Second, matched detections are used to compute precision-recall (PR) curves for each object class. The Average Precision (AP) is defined as the area under the PR curve, and the final metric, mean Average Precision (mAP), is the arithmetic mean of AP across all classes:

\begin{equation*}
    \text{mAP} = \frac{1}{N} \sum_{i=1}^{N} \text{AP}_i
\end{equation*}

where $N$ is the number of object classes ($4$ in our case) and $\text{AP}_i$ is the Average Precision for class $i$. This evaluation reflects both detection accuracy and localization quality across all classes and is among one of the most commonly used 3D detection metrics.


\subsection{Results on Synthetic Data}

As mentioned before, we created $100$ synthetic test samples to evaluate the performance of both model-based and deep learning detection methods on the \textit{same} data. Table \ref{tab:eval} presents the Mean Average Precision (mAP) and class-wise AP scores for both approaches. The SASA network was trained for 100 epochs on 80 training scenes containing 4518 labeled objects, while the model-based method required no training data and was only tuned on a small validation subset before being evaluated on the full test set.  

Overall, SASA achieves slightly superior performance with an mAP of $0.98$, compared to $0.97$ for the model-based method. The class-wise breakdown shows that both approaches perform consistently across all object categories, with SASA achieving marginally higher scores for \textit{reef\_cone} and \textit{tetrapod\_b}, while the model-based method is competitive for \textit{tetrapod\_s}. Importantly, both methods demonstrate robustness to object variability, which is crucial for generalizing to more complex underwater environments.  

Another noteworthy observation is the number of predictions: SASA produced $991$ detections compared to $957$ for the model-based method. This small difference suggests that SASA may capture slightly more difficult cases or overlapping objects, while the model-based method remains more conservative but highly reliable. The near-equivalent performance highlights an important trade-off: while SASA requires a large set of annotated training data and substantial computational resources, the model-based method achieves almost the same level of accuracy without any training data. This makes the latter especially attractive for practical scenarios where annotated underwater datasets are scarce or costly to acquire.

\begin{table}[tbp]
\centering
\caption{Performance Comparison on Synthetic Data (20 test scenes with 973 ground truth objects)}

\label{tab:eval}
\resizebox{\columnwidth}{!}{%
\begin{tabular}{lcccccccc}
\hline
\textbf{Model Type} & \textbf{mAP} & \textbf{Predictions} & \textbf{Reef\_Ring} & \textbf{Reef\_Cone} & \textbf{Tetrapod\_B} & \textbf{Tetrapod\_S} \\
\hline
SASA          & 0.98 & 991 & 0.98 & 0.99 & 0.98 & 0.96 \\
Model-Based   & 0.97 & 957 & 0.98 & 0.98 & 0.97 & 0.97 \\
\hline
\end{tabular}%
}

\end{table}

\subsection{Results on Sonar Data}

The results on synthetic data demonstrated that both approaches can achieve high performance under controlled conditions. We now turn to the main goal of this work: detecting objects in real-world sonar data. For this evaluation, we used the 15 annotated sonar test scenes containing 373 objects, as described in Section~\ref{sec:dataset}. The SASA network was applied directly without retraining, using the weights obtained from the synthetic dataset (80 training samples). For the model-based approach, only minor adjustments were required, specifically to the segmentation parameters and \gls{icp} thresholds, to account for the varying point density and noise levels in the sonar data.  

As shown in Table~\ref{tab:eval_sonar}, the model-based approach significantly outperforms the deep learning method, achieving a mean Average Precision (\gls{map}) of 0.83 compared to 0.40 for SASA. This large performance gap indicates that the domain gap between synthetic and real sonar data is substantial and that the learned features of SASA do not generalize well without domain adaptation. In contrast, the model-based approach, which relies directly on geometric alignment, remains more robust to the sensor-specific characteristics and noise patterns.  

Looking at class-wise AP scores, the model-based method performs consistently across most object categories. The largest drop is observed for \textit{reef\_ring} objects, which are often heavily cluttered or stacked in piles. This difficulty is primarily caused by segmentation errors rather than deficiencies in the detection stage: when multiple rings overlap or occlude one another, the sliding window segmentation tends to merge them into ambiguous clusters, which limits subsequent detection accuracy.

Whereas the deep learning-based SASA network showed much more pronounced variance across object classes. For example, SASA achieved performance comparable to the model-based approach for \textit{tetrapod\_b} objects (0.85 AP) and a reasonable score for \textit{tetrapod\_s}. In contrast, its performance dropped sharply for \textit{reef\_ring} objects, with only 0.04 AP. This suggests that the network is more effective at detecting objects that are well separated from the seabed. Put simply, SASA is better at identifying objects with larger vertical dimensions (e.g., \textit{tetrapod\_b} at 2.08 meters) than low-profile objects such as \textit{reef\_ring} (0.75 meters in height).  

\begin{figure*}[tbp]
    
    \captionsetup{justification=centering}
    \centering
    \subfloat[Ground Truth]{
        
        \begin{tabular}{@{}c@{\hspace{0.5em}}c@{\hspace{0.5em}}c@{}}
            \includegraphics[width=0.33\linewidth, trim={15cm 1cm 15cm 0cm}, clip]{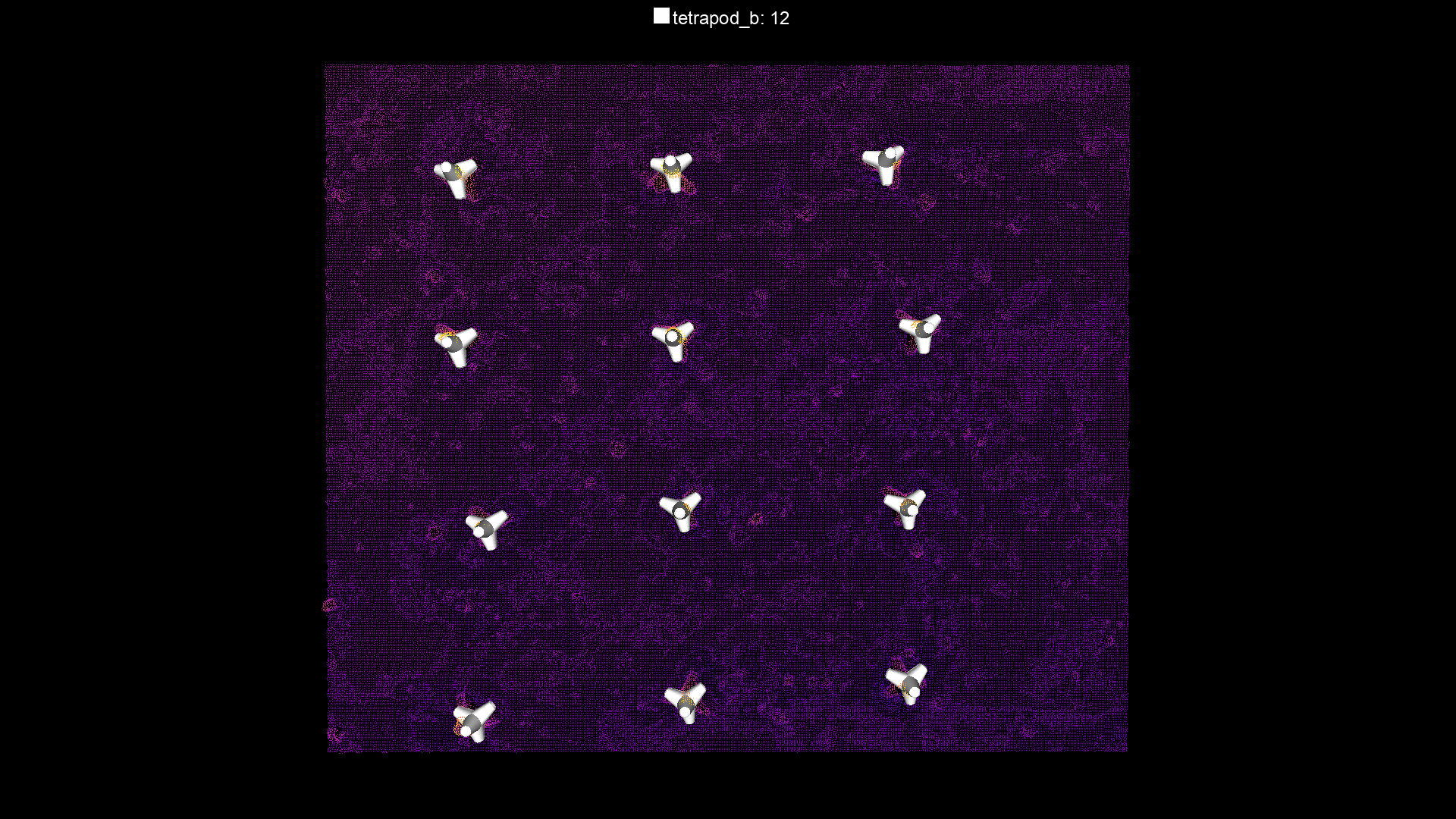} &
            \includegraphics[width=0.33\linewidth, trim={15cm 1cm 15cm 0cm}, clip]{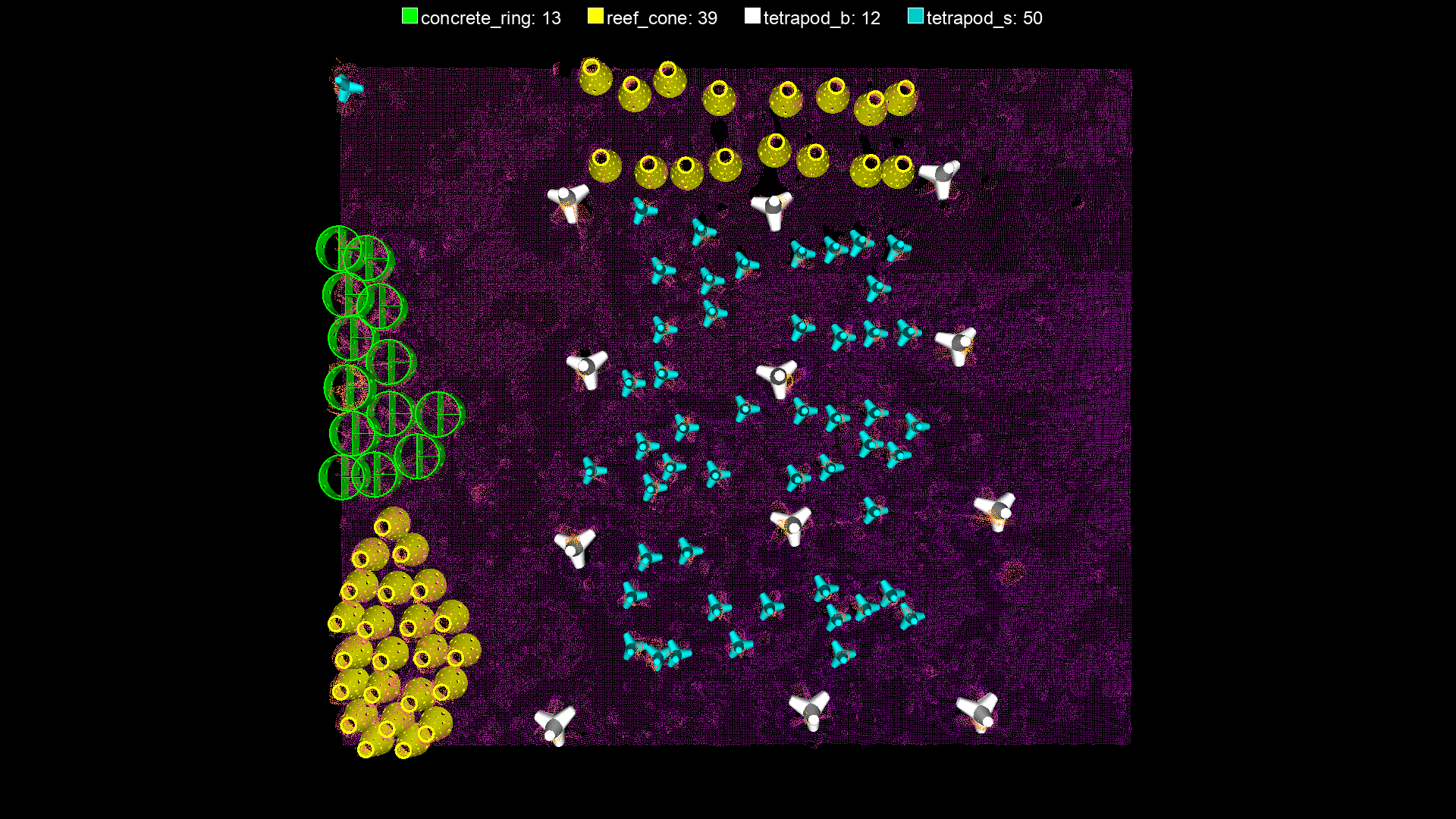} &
            \includegraphics[width=0.33\linewidth, trim={15cm 1cm 15cm 0cm}, clip]{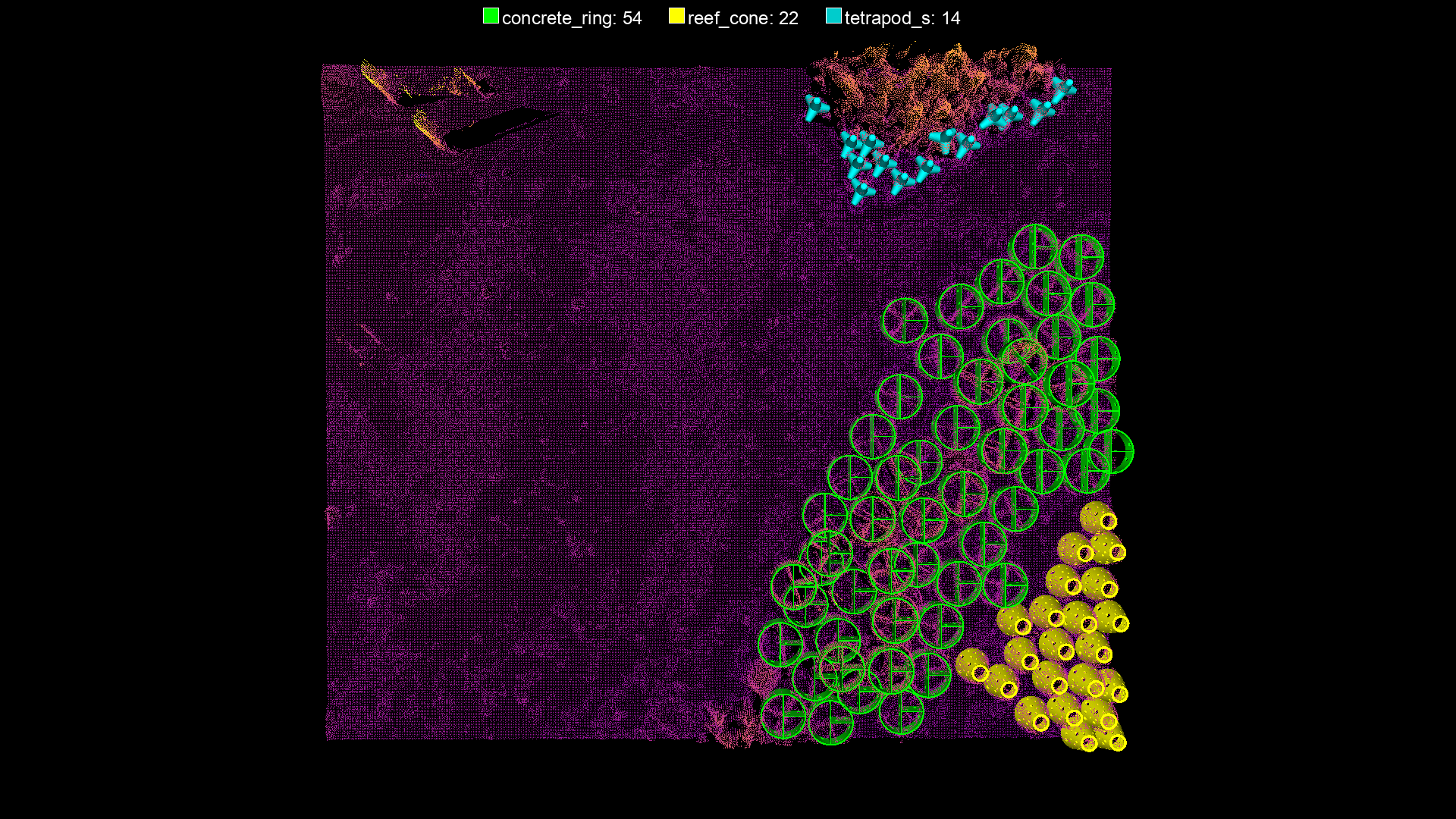} \\
        \end{tabular}
    }
    
    
    \subfloat[Predictions: Model Fitting]{
        \begin{tabular}{@{}c@{\hspace{0.5em}}c@{\hspace{0.5em}}c@{}}
            \includegraphics[width=0.33\linewidth, trim={15cm 1cm 15cm 0cm}, clip]{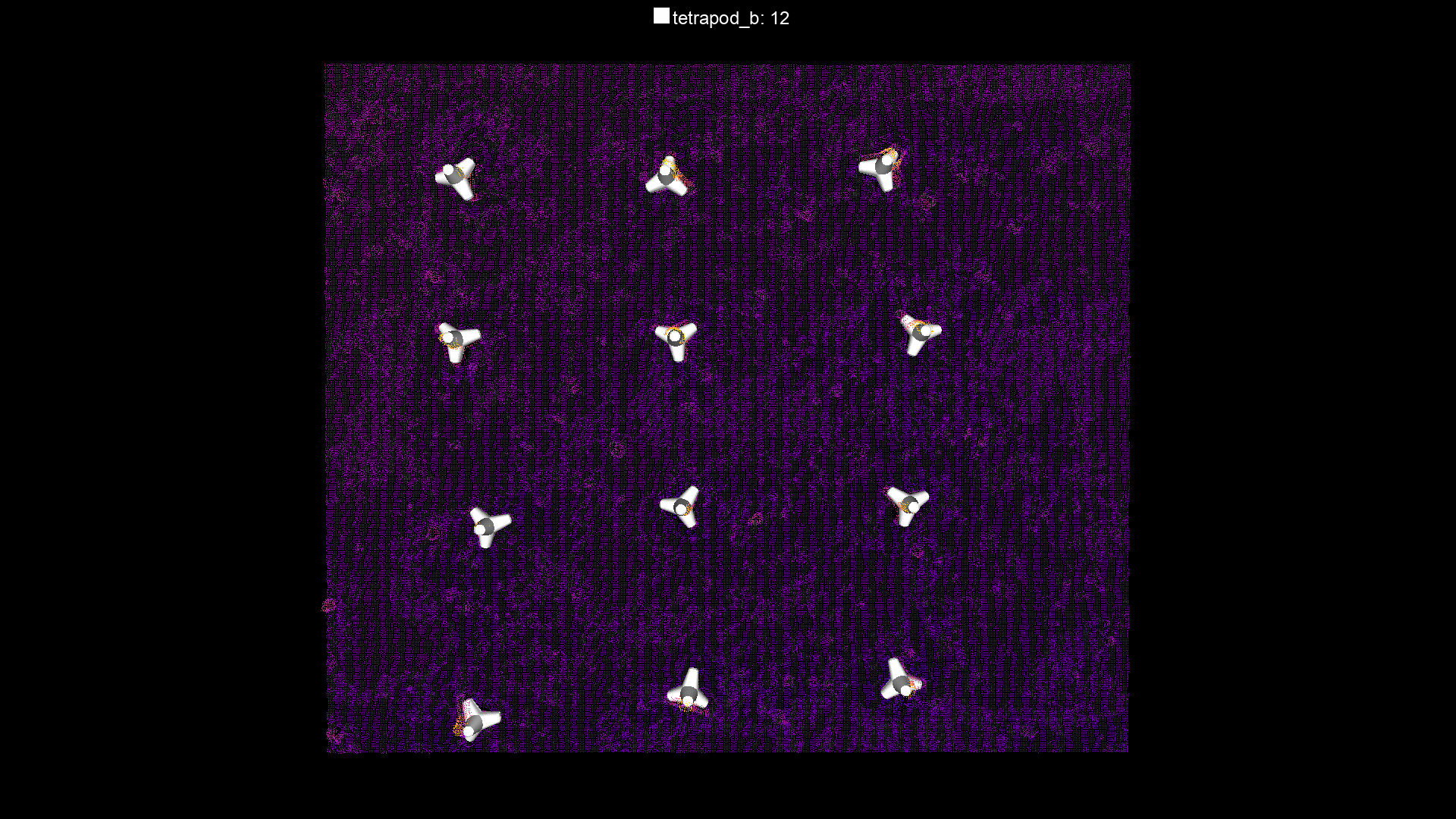} &
            \includegraphics[width=0.33\linewidth, trim={15cm 1cm 15cm 0cm}, clip]{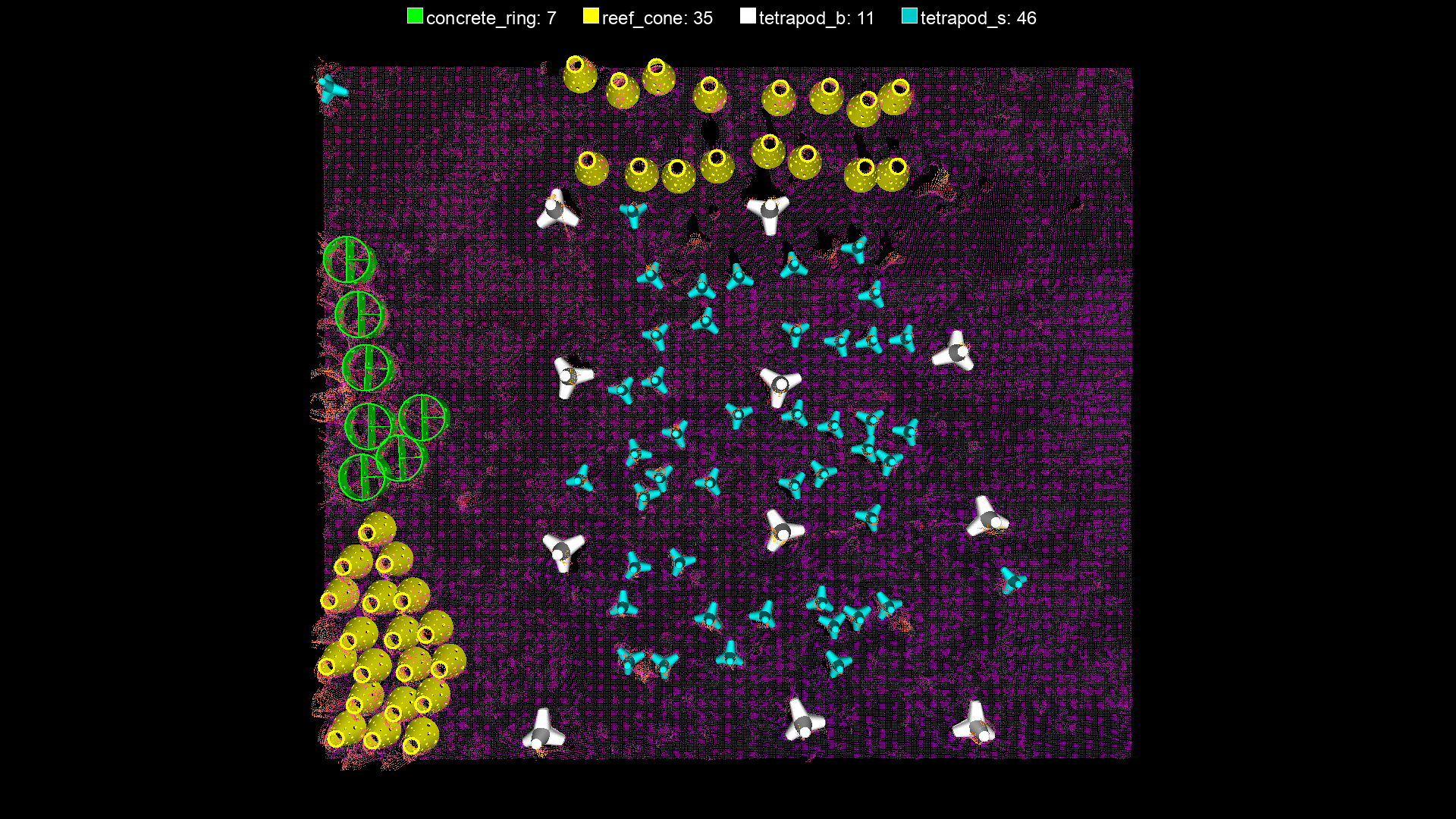} &
            \includegraphics[width=0.33\linewidth, trim={15cm 1cm 15cm 0cm}, clip]{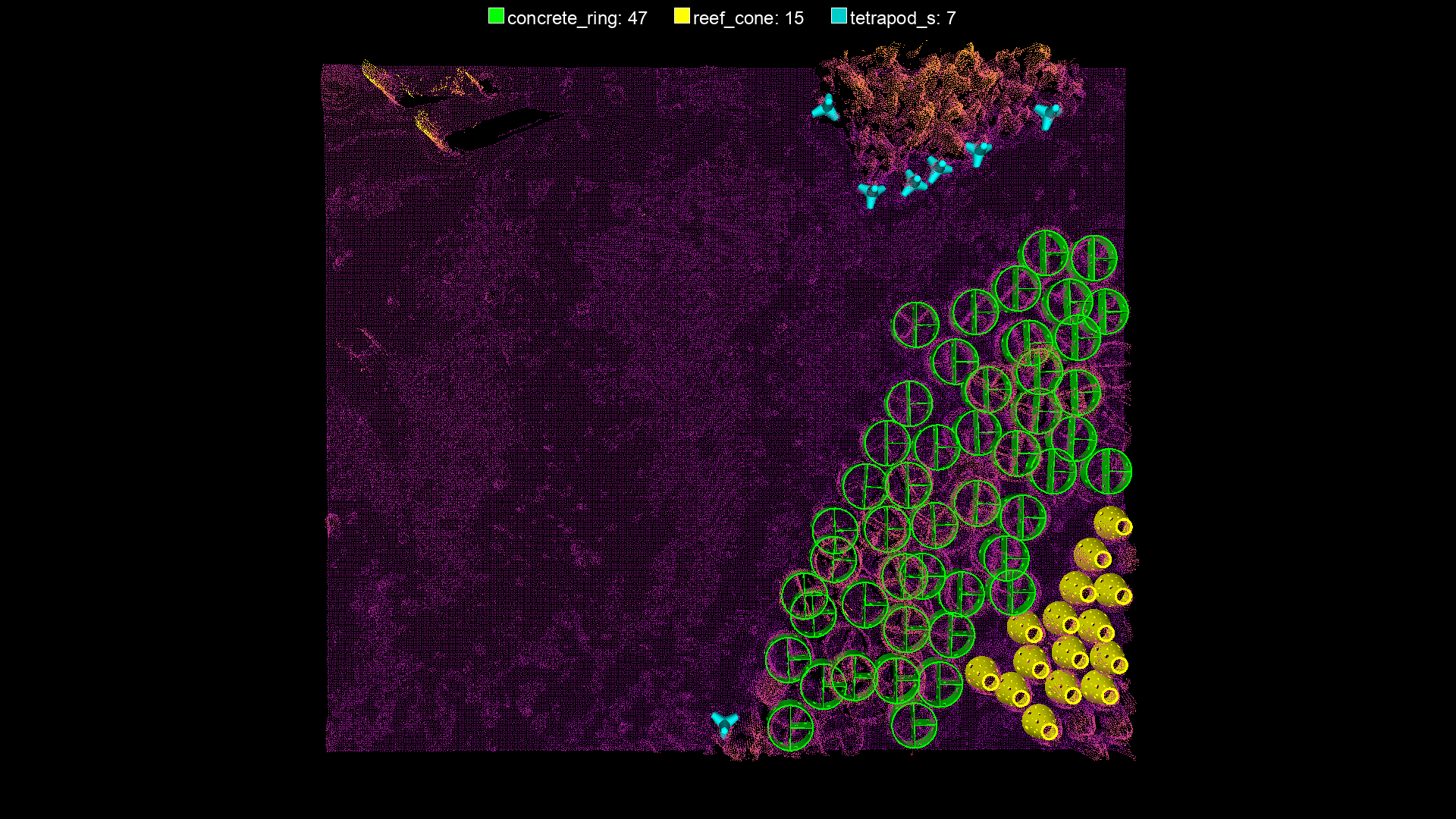} \\
        \end{tabular}
    }
    
    
    \subfloat[Predictions: SASA]{
        \begin{tabular}{@{}c@{\hspace{0.5em}}c@{\hspace{0.5em}}c@{}}
            \includegraphics[width=0.33\linewidth, trim={15cm 1cm 15cm 0cm}, clip]{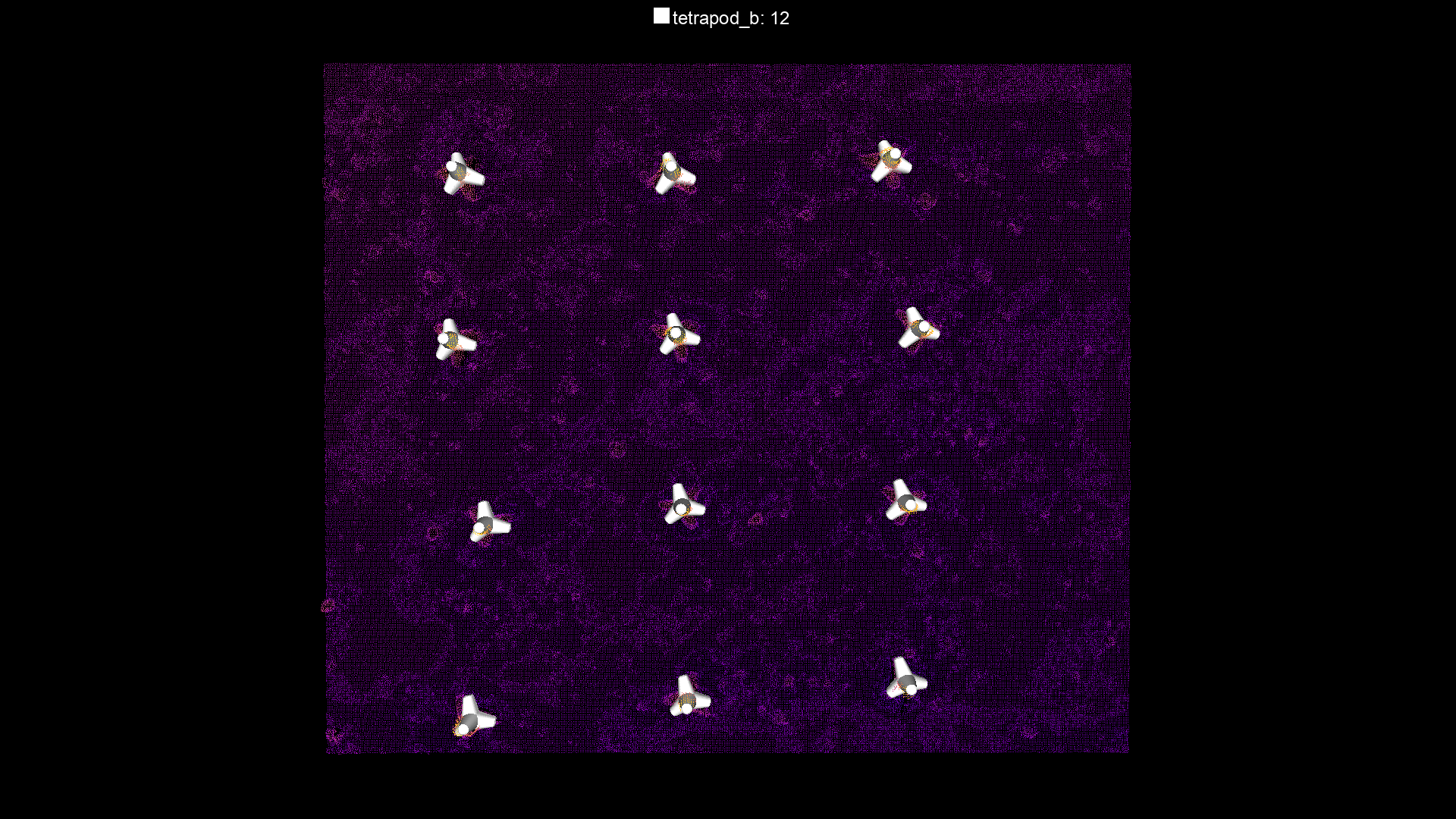} &
            \includegraphics[width=0.33\linewidth, trim={15cm 1cm 15cm 0cm}, clip]{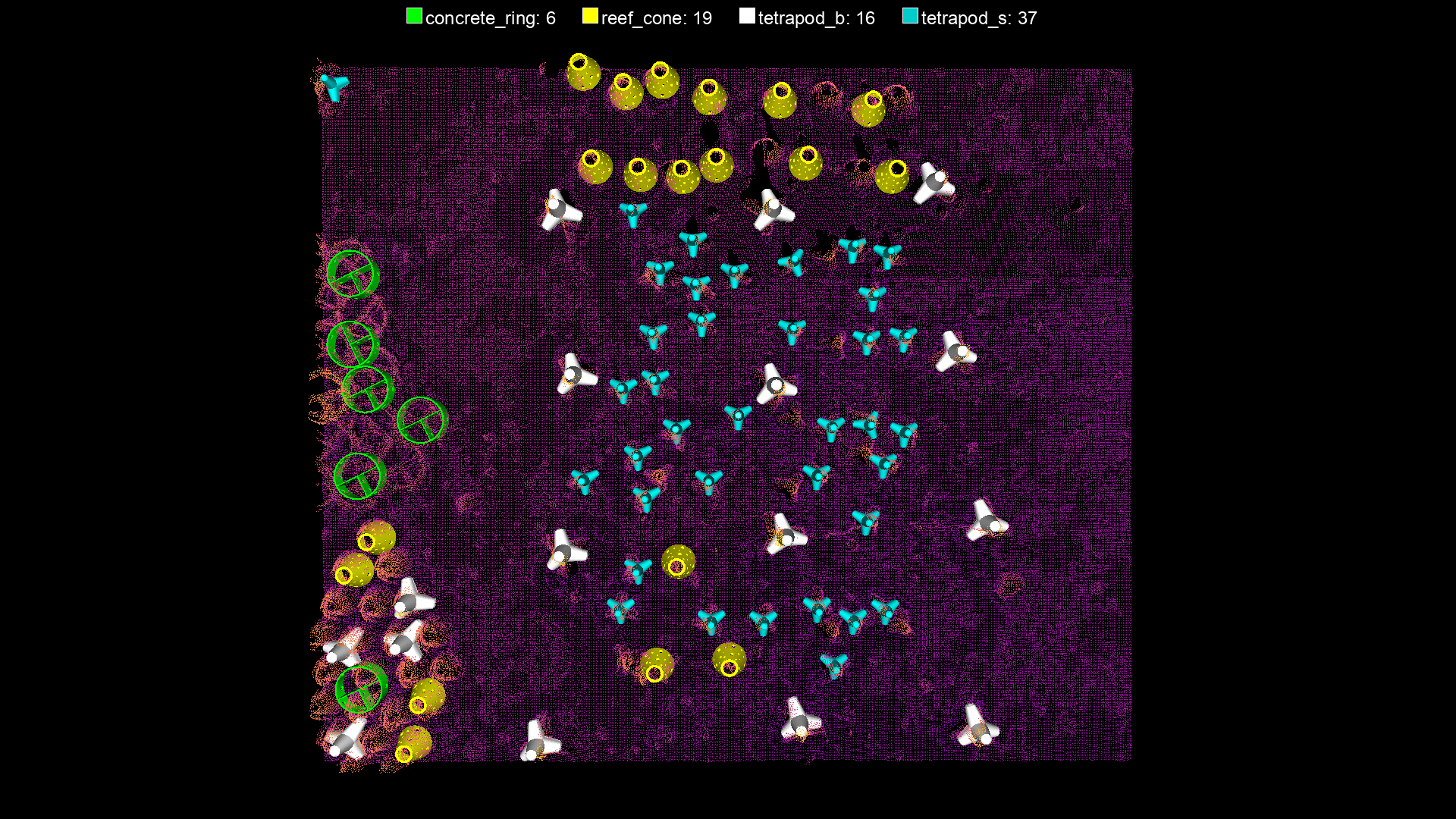} &
            \includegraphics[width=0.33\linewidth, trim={15cm 1cm 15cm 0cm}, clip]{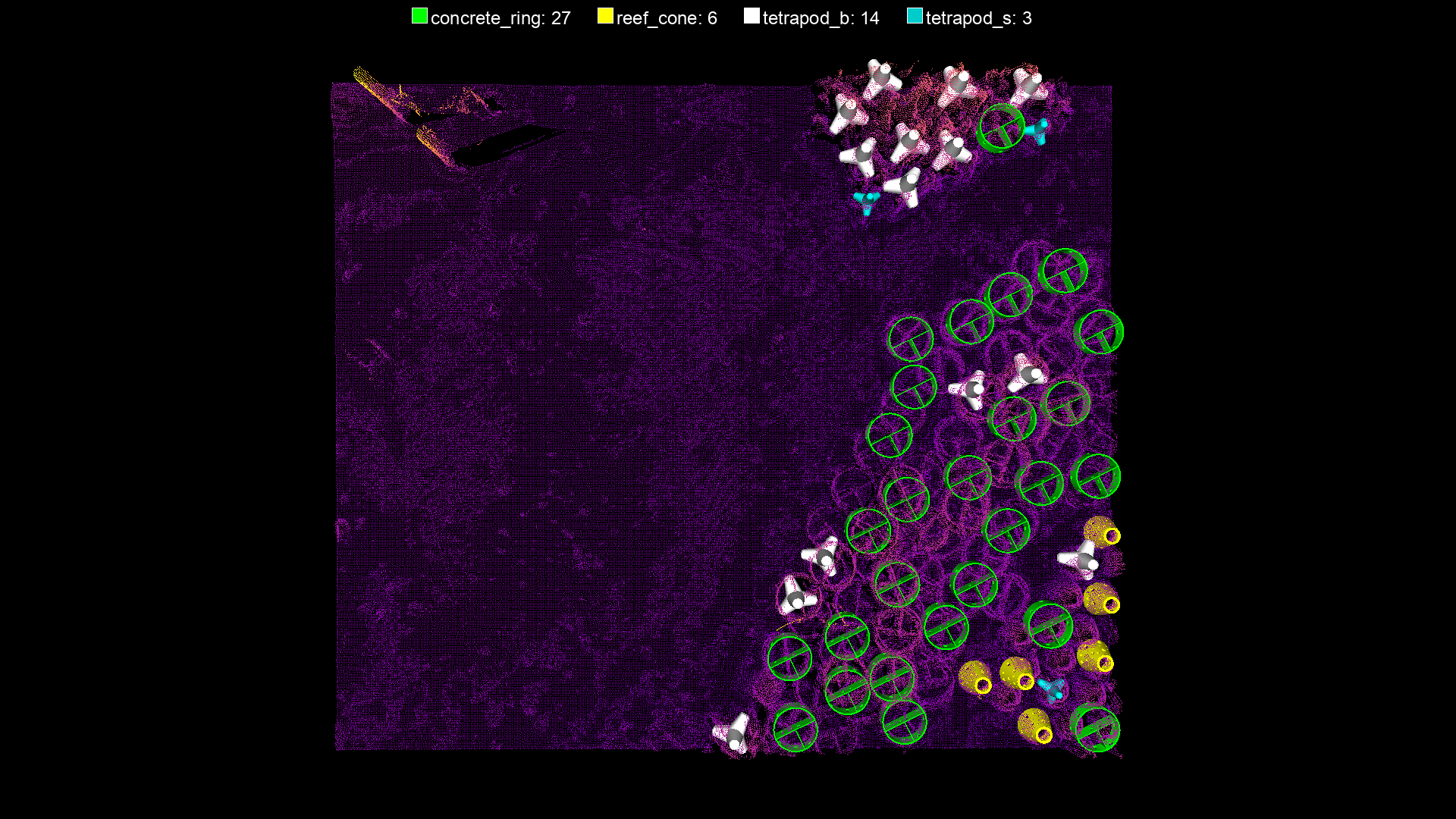} \\
        \end{tabular}
    }
    \caption{Sonar Detection Results: Comparison of ground truth, model-fitting, and SASA results for sonar point clouds detection. The predictions are shown by the colored 3D mesh-models of each object placed on the sonar data with detected position and pose.}
    \label{fig:3d_scenes_comparison}
\end{figure*}

Qualitative results are shown in Figure~\ref{fig:3d_scenes_comparison}, where predictions from both approaches are visualized alongside ground truth annotations for three sonar scenes. For clarity, 3D mesh models are displayed instead of bounding boxes. As seen, both approaches perform equally well on the simple scene in the first column containing 12 well-separated \textit{tetrapod\_b} objects. In more complex cases, the model-based approach largely maintains its accuracy with few exceptions, while SASA performs well when objects are spatially separated but fails in cluttered scenes (e.g., \textit{reef\_cone}) or heavily piled configurations (e.g., \textit{reef\_ring} in the last column).  

The sharp decline in SASA’s performance on real sonar data does not reflect its capacity to learn the 3D detection task. As shown in Section~\ref{sec:results}, the network achieved excellent results on synthetic data when sufficient training was available. Instead, its poor generalization is due to the well-known domain shift between simulated and real sonar. We hypothesize that narrowing this gap by incorporating more realistic \gls{mbes} noise models and object placements that include clutter and piling will significantly improve deep learning performance. Hence, this will form an important direction for our future work.


\begin{table}[tbp]
\centering
\caption{Performance Comparison on Real Sonar Data (15 test scenes with 373 ground truth objects)}
\label{tab:eval_sonar}
\resizebox{\columnwidth}{!}{%
\begin{tabular}{lcccccccc}
\hline
Model Type & mAP & \textbf{Predictions} & Reef\_Ring & Reef\_Cone & Tetrapod\_B & Tetrapod\_S \\
\hline
Model-Based   & 0.83 & 324  & 0.72 & 0.84 & 0.94 & 0.80 \\
SASA          & 0.40 & 233  & 0.04 & 0.17 & 0.85 & 0.52 \\
\hline
\end{tabular}%
}

\end{table}

\subsection{Training Data Requirements}
Since, in this publication, our core focus is on detecting underwater objects in absence or shortage of training data, it is crucial to understand how the amount of training data impacts detection performance of the neural network. In other words,  we question: how much training data would be sufficient to directly use a neural-based approach and not investing efforts into tuning model-based approach or creating a simulation model.

Therefore, we performed additional experiments with varying training data percentages. For this purpose, we created sub-training datasets with 25, 50, and 100 percent of the available 80 training scenes. While keeping the test dataset same as before with 20 scene samples.
Table \ref{tab:eval} shows the achieved \gls{map} for different training configurations on both synthetic and sonar data. Several key observations emerge from this analysis. First, even with only 25\% of the training data, SASA achieves performance equal to the model-fitting based approach on synthetic data (0.96 mAP).  As for the sonar data, the performance of the SASA network drops as the amount of training data is reduced. 
The most significant insight from these experiments is revealing. We still need approximately 1000 object annotations for the neural network to match the model-based approach performance. This indicates that substantial labeled data is still required to match model-based performance.

\begin{table}[b]
\centering
\caption{Average Precision evaluation results on synthetic data with varying training data percentages}
\label{tab:sub_eval}
\resizebox{\columnwidth}{!}{%
\begin{tabular}{lccccc}
\hline
Model Type & Training Percentage & Training Scenes & Training Annotations & mAP (Synthetic) & mAP (Sonar) \\
\hline
SASA      & Full dataset        & 80     & 4518   & 0.98 & 0.40 \\

SASA      & 50\% of data        & 40     & 2372    & 0.97 & 0.27 \\

SASA      & 25\% of data        & 20     & 1274    & 0.96 & 0.20 \\
\addlinespace[0.5em]
Model-Based       & No training required & ---    & ---      & 0.97 & 0.83 \\
\hline
\end{tabular}%
}
\end{table}

\section{Conclusion \& Future Work}\label{sec:discussion}
In this work, we set out to design underwater 3D object detection techniques tailored for large-area seafloor surveys, with the goal of avoiding costly and time-consuming annotated training data. To this end, we investigated two complementary approaches: a deep-learning method based on the state-of-the-art SASA network, trained exclusively on synthetic sonar data, and a traditional model-based method that relies on template matching without training data.

Our contributions are threefold: (i) developing a procedural \gls{mbes} simulation framework to generate synthetic sonar datasets, together with the implementation of a deep-learning based detection method trained exclusively on this simulated data, (ii) the implementation of a training-less, model-based detection pipeline, and (iii) the first large-scale evaluation and direct comparison of model-based and deep-learning based approaches for multi-class 3D object detection on sonar point clouds without using any real-world training data.

Results revealed a sharp contrast between methods. On synthetic data, the SASA network slightly outperformed the model-based approach, demonstrating the potential of learning-based methods when sufficient labeled data is available. On real sonar data, however, the model-based pipeline showed a clear advantage, achieving more than double the detection performance of SASA. This highlights a key insight: while deep learning excels in controlled synthetic settings, model-based techniques currently remain more robust when confronted with the noise, variability, and clutter of real sonar surveys.

We also addressed an important practical question: how much annotated sonar data would be required for deep learning to match model-based performance? Our findings suggest that a substantial volume of labeled data would be necessary, underlining the value of simulation and training-less methods in data-scarce domains.

Looking ahead, future work will focus on several directions aimed at bridging the gap between synthetic and real-world sonar data. First, we plan to further improve the \gls{mbes} simulation framework, particularly by enhancing noise modeling, incorporating environmental effects such as salinity and temperature variations, and developing more accurate sensor response models. Second, the realism of synthetic datasets will be extended by modeling natural phenomena, such as objects partially sinking into seabed, bio-fouling over time, and the presence of diverse natural clutter including rocks and irregular seabed textures. We believe these additions will help reduce the domain gap between simulation and reality, thereby improving the robustness of deep-learning based approaches.

In conclusion, our findings demonstrate that training-less 3D underwater object detection is both feasible and effective, establishing a solid baseline for future research. More broadly, this work points toward a scalable path for automated, training-free seafloor mapping, including robust object detection in challenging underwater environments.


%

\section*{Acknowledgments}
This research is funded by the German Federal Ministry of Education and Research (BMBF) as a part of the OTC-Digital Twin \& Analytics (OTC-DaTA) project (03ZU1107FB). The authors gratefully acknowledge Teledyne Marine for providing the Multibeam Echo Sounder (MBES) sonar data essential to this research.

\appendix[MBES  Sonar Noise Approximation]
\label{appendix}
To characterize \gls{mbes} measurement noise for simulation, we used a flat algae table surface from our survey data as a controlled reference with known ground truth geometry of a plane. This approach isolates sensor measurement uncertainty from natural seafloor roughness variations. To approximate the plane, we used RANSAC plane fitting and measured orthogonal distances from each measurement point to the fitted plane. These point-to-plane distances represent the total measurement error combining sensor noise and surface artifacts.

\begin{figure}[tbp]
\centering
\includegraphics[width=\columnwidth]{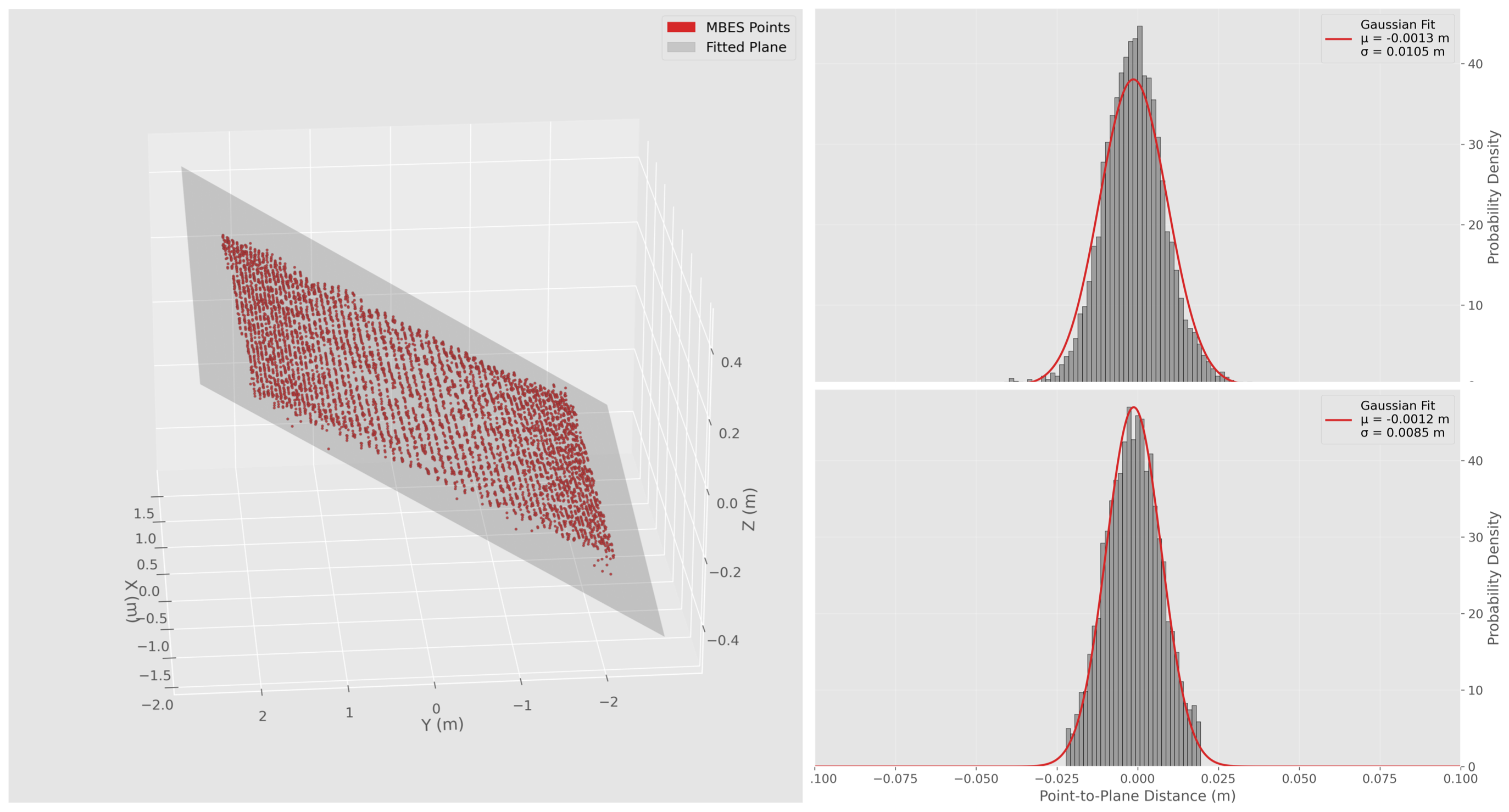}
\caption{MBES noise characterization using algae table as the reference surface. Left: 3D visualization showing RANSAC-fitted plane and MBES measurement points. Right: Point-to-plane distance distributions showing raw measurements (top) and after statistical outlier removal using Z-score.}
\label{fig:algae_analysis}
\end{figure}

Initial analysis of the raw distribution ($\mu = -0.001$ m, $\sigma = 0.010$ m) showed deviations from perfect normality in statistical tests. However, applying statistical trimming (removing points with Z-scores $> \pm 2$) improved the results, with the skewness normality test \cite{d1990suggestion} passing after outlier removal.

Figure \ref{fig:algae_analysis} shows the measurement distributions before and after outlier removal, along with the 3D plane fitting visualization. The removed outliers (statistical trimming) could likely represent surface artifacts such as algae growth though the exact nature of these artifacts cannot be definitively determined from the acoustic data alone.

The trimmed dataset exhibited near-zero mean ($\mu \approx -0.001$ m) and standard deviation ($\sigma \approx 0.008 $ m) with approximately Gaussian characteristics. For simulation purposes, we used a simplified Gaussian noise model $\mathcal{N}(0, 0.01^2)$ to approximate MBES measurement uncertainty.

\bibliography{bibliography}

\end{document}